\documentclass{svproc}

\usepackage{graphicx}
\usepackage{multirow}
\newcommand\tab[1][0.3cm]{\hspace*{#1}}
\usepackage{lscape}

\begin{document}
\mainmatter 
\title{Behaviour Trees for Creating Conversational Explanation Experiences}

\titlerunning{Behaviour Trees for Conversational Explanation Experiences}

\author{Anjana Wijekoon\and David Corsar\and Nirmalie Wiratunga}

\authorrunning{Anjana Wijekoon et al.}

\tocauthor{Anjana Wijekoon, David Corsar, Nirmalie Wiratunga}
\institute{School of Computing, Robert Gordon University, Aberdeen, Scotland\\
\email{a.wijekoon1@rgu.ac.uk}}

\maketitle

\begin{abstract}
Interactive multi-shot explanation experiences are key to achieving user satisfaction and improved trust in using AI for decision-making. This paper presented an XAI system specification and an interactive dialogue model to facilitate the creation of these explanation experiences. Such specifications combine the knowledge of XAI, domain and system experts of a use case to formalise target user groups and their explanation needs and to implement explanation strategies to address those needs. Formalising the XAI system promotes the reuse of existing explainers and known explanation needs that can be refined and evolved over time using evaluation feedback from the users. The abstract EE dialogue model formalised the interactions between a user and an XAI system. The resulting EE conversational chatbot is personalised to an XAI system at run-time using the knowledge captured in its XAI system specification. 
This seamless integration is enabled by using BTs to conceptualise both the EE dialogue model and the explanation strategies. 
In the evaluation, we discussed several desirable properties of using BTs over traditionally used STMs or FSMs. BTs promote reusability of dialogue components through the hierarchical nature of the design.   
Sub-trees are modular, i.e. a sub-tree is responsible for a specific behaviour, which can be designed in different levels of granularity to improve human interpretability.
The EE dialogue model consists of abstract behaviours needed to capture EE, accordingly, it can be implemented as a conversational, graphical or text-based interface which caters to different domains and users. 
There is a significant computational cost when using BTs for modelling dialogue, which we mitigate by using memory. Overall, we find that the ability to dynamically create robust conversational pathways makes BTs a good candidate for designing and implementing conversation for creating explanation experiences. 

\end{abstract}

\section{Introduction}
\label{sec:intro}

Today Artificial Intelligence~(AI) is being used by policymakers and industries to automate decisions that affect the general public. 
Advances in AI have impacted the transparency of these systems, making their decision-making process opaque. 
Explainable AI~(XAI) research has created a myriad of tools~(i.e. explainers) for explaining these AI systems and their decisions~\cite{arrieta2020explainable}. These integrated with an AI system are expected to meet the demands of GDPR right-to-explanation~\cite{sovrano2020making} and policies promoted by DARPA~\cite{gunning2019darpa}. However, often these explainers are used in isolation where an AI system extends to an XAI system by being able to provide a one-size-fits-all explanation using a single explainer for all users. Unlike the empirical precision of an AI system, an XAI system performance is measured by user-centred metrics like user satisfaction and improved trust~\cite{hoffman2018metrics}. Recent studies have found that these one-shot XAI systems need to evolve toward interactive multi-shot systems using a collection of explainers that are interactive, and iterative to address different user needs and to provide intelligible explanations~\cite{gunning2019darpa,dhanorkar2021needs}. 

This paper presents an interactive dialogue model that can be personalised for a given use case to create rich explanation experiences for users. An XAI system specification is proposed to formalise the system which captures the knowledge of domain, system and XAI experts. It consists of three key dimensions: AI system and users, Explanation Strategy, and Evaluation Strategy. The AI system and users are formalised by system and domain experts, which describe the AI model, user groups~(i.e. personas), and explanation needs of each user group. An XAI expert together with the system and domain experts formalises an Explanation Strategy which is a constellation of explainers curated to address explanation needs identified in the first dimension. Behaviour Trees~(BTs) are used in the conceptual modelling of the Explanation Strategies. The Evaluation Strategy is formulated by the XAI, system and domain experts considering what are the goals of the XAI system and what metrics are suitable to measure to see if they are met. This often takes the form of a questionnaire, which measures goals such as improved trust, user satisfaction and improved efficiency~\cite{hoffman2018metrics}. 

Once the XAI system specification is implemented, the XAI system is ready to provide explanations to users that meet their explanation needs. 
We view any interactions users have that expose them to explanations which impact their future interactions with the system as Explanation Experiences~(EE). 
We view these interactions to be conversational and through an XAI chatbot, we create EEs.
Accordingly, we define an EE as an episode of interactions of a user with the EE chatbot, such as expressing their explanation needs, receiving explanations, having arguments and evaluating the interactions.
The conversation is based on a dialogue model formalised using BTs referred to as the EE dialogue model. The EE dialogue model is designed to be abstract, such that each behaviour is open to interpretation by different domains, use cases and interaction modalities. By using BTs as the conceptual modelling tool for dialogue modelling and explanation strategies EE dialogue model enables dynamic personalisation and changes to conversational pathways based on an XAI system specification. The EE dialogue model also inherits many other desirable properties from BTs such as reduced susceptibility to error, reduced complexity, human interpretability, reusability of components, and reduced implementation cost compared to other tools like state transition models or finite state machines. 

\noindent The contributions of this paper are as follows. 
\begin{itemize}
    \item an XAI system specification that captures three key dimensions required to implement explainability within an AI system; it is evaluated for completeness by instantiating for three real-world use cases: loan approval, Radiograph fracture detection and recidivism prediction;
    \item a dialogue model that captures key interactions of users with the XAI system formalised using BTs; it is validated by dynamically personalising it for the Radiograph fracture detection use case and by creating an explanation experience of a clinician using the EE chatbot; and
    \item an evaluation of properties of the EE dialogue model that lends to creating rich explanation experiences with the users.
\end{itemize}

The rest of the paper is organised as follows. Section~\ref{sec:related} reviews literature in three related research areas, namely, the need for interactive multi-shot XAI, conceptual modelling of dialogue and behaviour trees. Section~\ref{sec:xai} presents the XAI system specification and evaluation with three use cases. Conversational requirements identified for the EE chatbot are presented in Section~\ref{sec:conversation}. Section~\ref{sec:bt} presents the BT fundamentals and our EE dialogue model as a BT. In Section~\ref{sec:discuss} we discuss the properties of the EE dialogue model that enable rich explanation experiences through conversation. Finally, conclusions and future directions are presented in Section~\ref{sec:conc}.
\section{Related Work}
\label{sec:related}

\subsection{Need for interactive multi-shot XAI}
Explainable AI~(XAI) is a mature field of study where the goal is to provide insights into how a black-box AI model works. 
These insights can be explaining how the model works, why a certain decision was made or what generally impacts model decisions~\cite{arrieta2020explainable}. Explanations should cater to different types of users~(i.e. personas), and their explanation needs~\cite{vilone2021notions,liao2020questioning}. 
We refer to an AI system which is capable of catering to these user needs as an XAI system. 
All facets of XAI lead to the same conclusion that XAI is not one-shot, meaning an XAI system should be interactive, able to provide multiple explanations considering different personas and their needs during design~\cite{sokol2020one,liao2020questioning}. 
We refer to a constellation of explainers curated to address such needs within an XAI system as an Explanation Strategy. 

Designing strategies and interactive interfaces to address the needs of many is an upcoming research domain within XAI~\cite{sokol2020one,schoonderwoerd2021human}. 
Authors of~\cite{sokol2020one} designed two XAI strategies one for explaining the model and one for explaining model decisions; each strategy consists of two or more explainers. 
These strategies were curated by the researchers to suit a wide range of domains and data types. 
In contrast, authors of~\cite{schoonderwoerd2021human} derive the XAI strategies from the domain experts~(clinical decision support for child health). 
A user study was conducted to learn user preferences, the results of which were used by researchers to curate XAI strategies for different explanation needs. 
Each resulting strategy was a combination of one or more explainers. Accordingly, there are both user-driven and expert knowledge-driven methods to curate XAI strategies. 
The usability of these patterns also relies on the interactive interface that converses with the user to understand their needs. Many authors used argumentation as as framework to model these interactions~\cite{baumann2021choices,cayrol2005acceptability,mcburney2002games,madumal2019grounded} which enables scrutability. 
Current literature considers the XAI strategies and the interaction models to be disjoint. Also, the lack of shared conceptual modelling across these two components makes them harder to integrate and reuse in different domains. 
In this paper, we propose a Behaviour Trees~(BT) as the conceptual model to design XAI strategies and interactions to improve implementation efficiency and reusability.

\subsection{Conceptual Modelling of Dialogue}
Conversation has been considered as a medium to implement interactive XAI in contrast to graphical or text-based user interfaces~\cite{sokol2020one,madumal2019grounded}. 
Conversational interactions of XAI may include how to understand user's explanation needs, presenting explanations of different modalities and getting user feedback. 
In general, conversational interactions are formalised as dialogue models using Argumentation Frameworks such as AAF~\cite{baumann2021choices}, BAF~\cite{cayrol2005acceptability} or ADF~\cite{mcburney2002games}.
Alternatively, dialogue models are graphically represented using State Transition Models~(STM)~\cite{hernandez2021conversational,madumal2019grounded} or Finite State Machines~(FSM)~\cite{le2018cognitive}. 
BTs are a less frequent choice for dialogue modelling; often they model robot interactions where dialogue refer to natural language instructions~\cite{bouchard2018modeling,suddrey2022learning,iovino2022survey}. 
In our work, we show BTs are preferred over other conceptual models like STM given its properties such as modularity, reusability and robustness to dynamic change. 

\subsection{Behaviour Trees}

BTs are conceptual structures that model behaviours that was introduced as an alternative to FSMs to improve modularity and reusability~\cite{colledanchise2018behavior,iovino2022survey}. 
A BT is directed rooted tree with a set of nodes that control the navigation and actions of an Actor~\cite{colledanchise2018behavior}. 
There are several properties of BTs that lends naturally to modelling dialogue. 
The hierarchical nature of BTs allows the design of a ``behaviour'' to be a sequence of sub behaviours. This also brings the ability to design a BT in different granularities, which means, any behaviour~(node) can be replaced by a sub-tree that describes it in higher granularity~\cite{florez2008dynamic}. 
The sequential nature of node execution enables the design of behaviours that depends on the successful completion of previous behaviours. 
Also, BTs are reactive, which means, any behaviour~(node execution) can be interrupted by a behaviour that takes higher priority. 
BTs are modular, meaning a node~(coarse granularity) or a sub-tree~(fine granularity) is responsible for a specific behaviour which lends to efficient design and implementation and promotes reusability. 

Literature has highlighted several other desirable properties of BTs over other methods of behaviour modelling that naturally advantages to dialogue modelling. In robotics, the designing of a BTs is either knowledge-driven or data-driven. Manually constructing a BT requires the domain experts to model the BT with suitable granularity using a BT editor which does not require any programming knowledge~\cite{iovino2022survey}. For example, authors of~\cite{bouchard2018modeling} model daily-living activities in a smart home using BTs. In contrast, the design can be data-driven; if the robot's expected behaviours are available as a corpus of transcriptions, BTs are generated by abstracting actions and navigations from the data~\cite{suddrey2022learning}. For example, when modelling a robot for cooking, recipes can provide information to extract abstract behaviours. 
Other desirable properties of BTs over alternatives include flexible design through ability to dynamically change in response to the environment~\cite{french2019learning,bouchard2018modeling}, expressiveness through granularity~\cite{bouchard2018modeling}, implementation efficiency through reusability~\cite{bouchard2018modeling,coronado2018development}, human interpretability~\cite{iovino2022survey}, less prone to error during development~\cite{coronado2018development} and reduced complexity~\cite{bouchard2018modeling,french2019learning}. 
In this paper, we design the BTs manually using conversational requirements extracted from domain knowledge and previous work~\cite{hernandez2021conversational,madumal2019grounded}. In addition, we demonstrate several desirable properties of BTs for modelling XAI interactions compared to STMs. 

BTs is a research field that is still evolving.
Accordingly, there is a lack of tools and frameworks that allow the design and execution of BTs. An analysis in 2019 showed that there are only about 10 libraries that are regularly maintained~(last update on or after 2018) and some of these are specific to behaviour modelling for games~(as a Unity3D plugin)~\cite{iovino2022survey}. 
This is a barrier to the widespread implementation of BTs in applications that are not game design. In our work, we use customised versions of Behavior3 libraries~\cite{marzinotto2014towards}~\footnote{https://github.com/behavior3} to design and execute BTs.

\section{XAI Systems}
\label{sec:xai}

An AI system consists of an AI model that automates a decision-making task for a specific set of users. Few example AI systems are presented in Table~\ref{tbl:examples}. In order to transform an AI system into an XAI system, it must be reinforced with additional information and techniques that enable the system to provide explanations to users, i.e. create explanation experiences~(EE). In this paper, such an XAI system is characterised using the XAI system specification. 

\begin{table}[ht]
\centering
\caption{Example AI systems}
\small
\renewcommand{\arraystretch}{1.2}
\begin{tabular} {p{7cm}p{5cm}}
\hline
AI System&Users\\
\hline
Loan approval system used by a mortgage provider that predicts if a loan application is approved or not&Loan applicants, Trainee loan officers, Regulatory bodies\\
\hline
Weather forecasting system that predicts the weather for the next 14 days using past weather data&General public, Meteorologists, Industries like aviation, transportation and agriculture\\
\hline
Radiograph fracture detection system that predicts if a fracture is present in a Radiograph&Radiologists, Clinicians, Patients, System administrators\\
\hline
Movie recommender system on a streaming platform that recommends the next movie to watch based on past preferences&General public, Streaming platform executives\\
\hline
Recidivism prediction system that predicts the likelihood of recidivism in the next two years of an inmate&Judges, Judicial review committees\\
\hline
\end{tabular}
\label{tbl:examples}
\end{table}

In order to facilitate creating EE with users, an XAI system should formalise necessary information related to its AI system, types of users, their explanation needs and algorithms that provide these explanations known as explanation strategies. In addition, we also highlight the need to evaluate the XAI system in order to improve the explanation strategy for future users. Accordingly, we characterise an XAI system using the following concepts categorised into three key topics: system description, explanation strategy, and evaluation strategy. These concepts are formalised using an ontology~\footnote{https://github.com/isee4xai/iSeeOnto}.

\subsection{Evaluating the XAI System Specification}
Once the XAI system specification is implemented, it can interact with the users to provide explanations. Accordingly, we evaluate the specification for completeness, i.e. whether it captures all required information to implement an XAI system that can create interactive explanation experiences. To evaluate the completeness, we instantiate the proposed XAI system specification using three use cases: Radiograph fracture detection system in Table~\ref{tbl:jiva}, loan approval system in Table~\ref{tbl:loan} and Recidivism prediction system in Table~\ref{tbl:recidivism}. Considering the paper length, we have included the loan approval system and recidivism prediction system specifications in Appendix~\ref{ap:xaispecs}.

Table~\ref{tbl:jiva} formalises the XAI system for a user group of clinicians who are using a Radiograph fracture detection system in their decision-making process. A clinician is seeking to understand why some Radiographs are marked as fracture and if the machine-generated explanations match their clinical knowledge~(by asking for other similar images). This design of XAI system specification is capable of capturing all required characterises and documenting the explanation and evaluation strategies designed for the clinician persona. Accordingly, the XAI system can be configured to provide interactive multi-shot explanations to any clinician who has similar needs. This will require configuring~(by reusing existing implementations) each explainer specified in the explainer strategy. The specification captures required parameters commonly used by explainers like AI model, data types, possible outcomes and the explanation target data instance. 

While completeness is achieved, we encountered that multiple specifications are required when the same use case is applicable to multiple personas with multiple explanation needs. The current design requires the documentation of specifications for each persona of the XAI system. For example, Table~\ref{tbl:jiva} is an XAI system described with respect to clinicians. Other personas interested in the Radiograph fracture detection system would be training Radiologists who use the system as a learning tool, patients who are trying to understand care pathways based on the system decision or system administrators who are looking to understand if the system affects clinician efficiency to reduce decision time. They have different questions, and to answer these questions they require different explanation strategies. Consequently, there will be different evaluation goals for each persona. In such a scenario, there will be multiple XAI system specifications where the AI system is repeated used. However, this is not a major overhead given there are only a limited number of personas~(see users in Table~\ref{tbl:examples}) for a given XAI system.  

\begin{landscape}
\begin{table}[ht]
\centering
\caption{XAI System Specification}
\small
\renewcommand{\arraystretch}{1.2}
\begin{tabular} {lllp{14cm}}
\hline
\multicolumn{3}{l}{\textbf{System Description}}&the characterises of the AI system formulated by the domain and system experts\\
&\multicolumn{2}{l}{\textit{AI System}}&the AI model used by the system to automate decisions\\
&&AI Task&the problem solved by the AI model\\
&&AI Method&the AI technique/algorithm that is used to solve the problem\\
&&Data&characterised by number of features, number of data points, and data types\\
&&Assessment&performance metrics of the AI Model\\

&\multicolumn{2}{l}{\textit{Persona}}&the characteristics of the user group\\
&&AI knowledge Level&measures the user's mental model with respect to AI knowledge\\
&&Domain Knowledge Level &measures the user's mental model with respect to domain knowledge\\
&&Resources&technical capacities of the edge device to receive explanations and interact\\

&\multicolumn{2}{l}{\textit{Explanation Need}}& known explanation needs of the persona to reuse\\
&&User questions&a collection of questions frequently asked by the user group~\cite{vilone2021notions}\\
&&intent&explanation need expressed by the questions asked by the users. Common intents are transparency, efficiency, trust and debugging~\cite{arrieta2020explainable}\\
&&Target&the entity that needs to be explained, this can be the model, a decision for a specific case or data.\\
\hline

\multicolumn{3}{l}{\textbf{Explanation Strategy}}&the constellation of explainers that are curated to address the explanation needs of the persona by an XAI expert together with domain and system experts. A strategy consists of one or more explainers, that can be used to answer known questions raised by the user group. A strategy is designed as a behaviour tree where each leaf action node is an explainer; navigation is determined by the indication of user satisfaction at each node.\\

\hline
\multicolumn{3}{l}{\textbf{Evaluation Strategy}}& a methodology to evaluate explanation experiences curated by the XAI, domoin and system experts~\cite{hoffman2018metrics}\\
&\multicolumn{2}{l}{\textit{Questionnaire}}& set of questions to ask the users to evaluate and understand the user perception of the experience\\
&\multicolumn{2}{l}{\textit{Response Interpretation Policy}}&a policy defined to evaluate the current explanation strategy works for the system or needs modification based on questionnaire responses\\
\hline
\end{tabular}
\label{tbl:xaispec}
\end{table}
\end{landscape}

\begin{landscape}
\begin{table}[ht]
\centering
\caption{XAI System: Radiograph fracture detection system for Clinicians}
\footnotesize
\renewcommand{\arraystretch}{1.2}
\begin{tabular} {lp{14cm}}
\hline
\textbf{System Description}\\
\textit{\tab AI System}\\
\tab \tab AI Task&Given a Radiograph, the system predicts if it contains a fracture or not\\
\tab \tab AI Method&A Convolutional Neural Network model performing binary classification\\
\tab \tab Data&AI Model was trained with 40561 data instances; each Radiograph is an image of 1500 x 2000 pixels. \\
\tab \tab Assessment& Accuracy of the model is 83.4\%\\

\textit{\tab Persona: clinician}\\
\tab \tab AI Knowledge Level&no knowledge~(from no knowledge, novice, advanced beginner, competent, proficient and expert)\\
\tab \tab Domain Knowledge Level&proficient~(from no knowledge, novice, advanced beginner, competent, proficient and expert)\\
\tab \tab Resources&Screen Display can handle touch and visual modalities\\
\textit{\tab Explanation Need}\\
\tab \tab User questions and Intents&Why is this Radiograph marked as "fracture"? $\rightarrow$ Transparency intent \\
&Are there similar Radiographs that are also marked as "fracture"? $\rightarrow$ Trust intent\\
\tab \tab Query&Radiograph of 1500 x 2000 pixels with outcome either "fracture" or "no fracture"\\
\hline
\textbf{Explanation Strategy}\\
\textit{\tab Explainers}&Feature attribution using Integrated Gradients and LIME to satisfy transparency intent\\
&Nearest neighbours to satisfy trust intent\\
\textit{\tab Behaviour Tree}&\includegraphics[width=.5\textwidth]{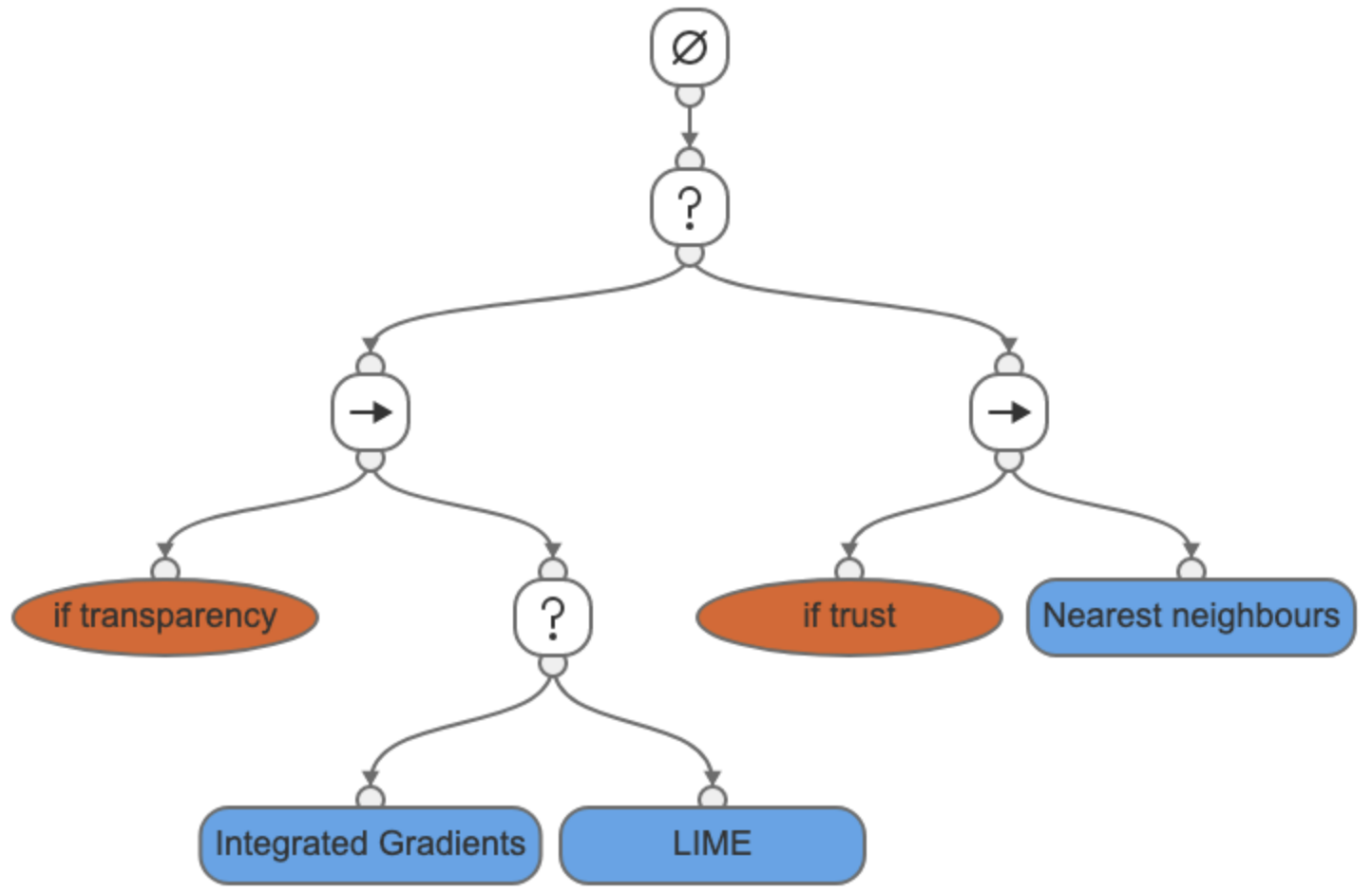}\\
\hline
\end{tabular}
\label{tbl:jiva}
\end{table}
\end{landscape}

\begin{landscape}
\begin{table}[ht]
\centering
\footnotesize
\renewcommand{\arraystretch}{1.2}
\begin{tabular} {lp{14cm}}
\hline
\textit{\tab Behaviour Tree Description}& If the user has transparency intent, first present a feature attribution explanation using Integrated Gradients, if the user is not satisfied, continue to present another feature attribution explanation using LIME. If the user has trust intent, present three nearest neighbour Radiographs from the training dataset. If the user is satisfied after receiving explanations for one intent exit, else, present explanations for the remaining intent. Exit after executing all three explanation techniques\\
\hline
\textbf{Evaluation Strategy}\\
\textit{\tab Questionnaire}&Question 1: The explanations that were presented had sufficient detail \par Answers: Likert scale, 5 items from Strongly Disagree to Strongly Agree\\
&Question 2: The explanations let me know how accurate or reliable the AI system is.\par Answers: Likert scale, 5 items from Strongly Disagree to Strongly Agree\\
&Question 3: The explanation lets me know how trustworthy the AI system is.\par Answers: Likert scale, 5 items from Strongly Disagree to Strongly Agree\\
\textit{\tab Interpretation Policy}&All three questions should receive positive feedback\\
\hline
\end{tabular}
\end{table}
\end{landscape}
\section{Conversational Requirements of an Explanation Experience}
\label{sec:conversation}

\begin{figure}[ht]
\centering
\includegraphics[width=\textwidth]{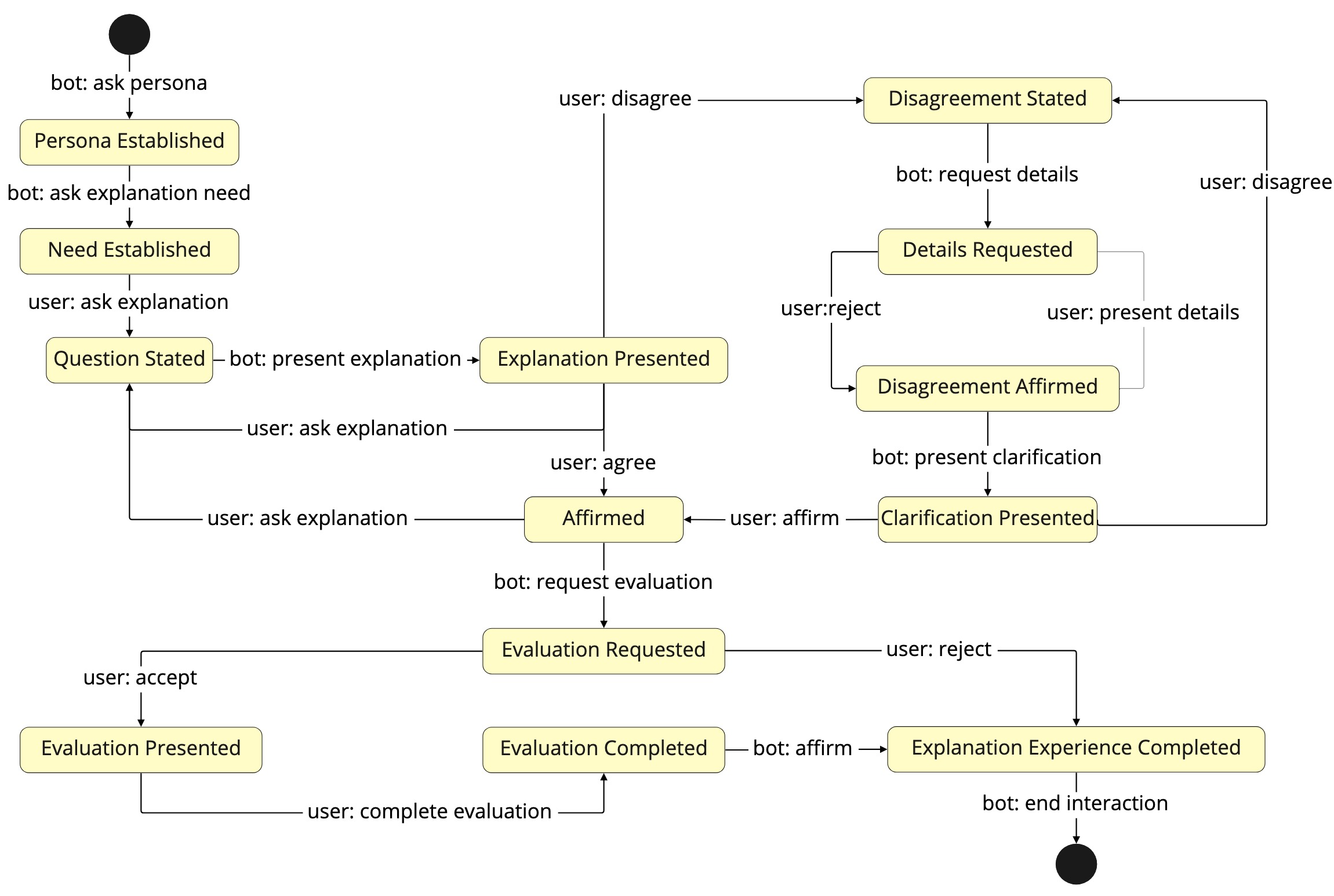}
\caption{State Transition Diagram}
\label{fig:fs_diagram}
\end{figure}

A rich and personalised Explanation Experience~(EE) can be captured through conversation~\cite{miller2019explanation}. Given an XAI system configured according to the specification, we imagine a chatbot within the XAI system that interacts with the end users. The chatbot will establish the persona and explanation needs to provide suitable explanations towards user satisfaction. Traditionally, the dialogue model of the EE chatbot can be designed as a state transition diagram. We use this technique to understand the conversational requirements of the EE Chatbot~(Figure~\ref{fig:fs_diagram}).

\subsection{Actors}

We imagine an explanation experience to take place between two parties. A \textit{user} who is seeking explanations and the EE \textit{chatbot}. The \textit{chatbot} establishes the persona and explanation needs of the user using question-answer type conversation. The \textit{user} can request an explanation by asking a question and accepting or disagreeing with an explanation received. Based on user questions, the \textit{chatbot} is able to present explanations and clarifications and record disagreements for future improvements. Finally, the \textit{user} has the opportunity to evaluate their experience using the questionnaire. 

\subsection{Understanding persona and explanation needs}
Characteristics of the persona such as AI and domain knowledge levels can be acquired in the form of question and answer. Resources can be derived from the edge device used by the \textit{user} to interact with the bot. 
Given that the domain and system experts have already defined a list of common questions and intents of the persona~(see Table~\ref{tbl:jiva}), the next step is to find out the exact need of the \textit{user}. Accordingly, to establish the explanation need, the \textit{chatbot} will present pre-configured questions to the \textit{user} to select one or allow them to pose their own question. Based on the question, the \textit{chatbot} will identify the intent. 
Based on the question/intent, the \textit{chatbot} will follow up with the \textit{user} to confirm the entity that needs to be explained. For example, in a Radiograph fracture detection system, the XAI system is already aware of the Radiograph the clinician is currently working with. 

\subsection{Executing the Explanation Strategy}
Once an intent is identified, the \textit{chatbot} finds the most suitable explainer from the explanation strategy~(note that the explainers in the strategy are linked to different intents, see Table~\ref{tbl:jiva}). The \textit{chatbot} executes the explainer with respect to the explanation target and other parameters provided, and the explanation is presented to the \textit{user}. Next, the \textit{user} has the opportunity to pose further questions or indicate their satisfaction. If the follow-up question is one of the pre-configured, the \textit{chatbot} will continue to select the most suitable explainer from the strategy to answer the question and this process will iterate until user is satisfied. In case user raises a question that cannot be answered by the explanation strategy, it will be recorded to improve the strategy for future users.

\subsection{Support for Argumentation}
The \textit{user} has the opportunity to indicate their disagreement with an explanation presented by the \textit{chatbot}. In such a scenario, the \textit{chatbot} will try to acquire further details from the \textit{user} to understand their mental model. Both the \textit{chatbot} and \textit{user} has opportunity to present supporting arguments and clarifications provided by the \textit{chatbot} can come from an explainer in the explanation strategy or from domain knowledge within the XAI system. 

\subsection{Evaluating Explanation Experience}
Finally, the \textit{user} has the opportunity to evaluate their experience. The \textit{chatbot} presents them with a pre-configured questionnaire to answer. Once a significant number of users who identify as the persona have created explanation experiences~(i.e. interacted with the chatbot, received multi-shot explanations and evaluated their experience), the XAI system can collate these evaluations of experiences using the results interpretation policy defined in the XAI system specification. This overall outcome, and any feedback received during conversations are used to refine the explanation strategy towards improved user experience. 
\section{Dialogue Model as a Behaviour Tree}
\label{sec:bt}

Implementing conversational systems based on STMs pose several challenges. The rigid nature of transitions makes the conversation less interactive and restricted to only several conversational pathways. In addition, any changes to the dialogue design results in major implementation changes. Accordingly, we propose to conceptualise the dialogue model using Behaviour Trees~(BT).

\subsection{Behaviour Trees}
\label{sec:btbasics}
BTs are a type of conceptual models that formalises the behaviours of an actor in a given environment. The tree structure is made of different types of nodes that implement behaviours and navigation. Each node has a state that indicates if the execution of the node was a success or failure~(indicated by the coloured border; in green for success and red for failure). There are two types of nodes that control navigation, also known as composite nodes: Sequence and Priority. The leaf nodes that implement a specific behaviours are known as Action Nodes. There are also decorator nodes and condition nodes to control access to a sub-tree. Importantly, each action node can be replaced by a sub-tree that represents the behaviour in finer granularity. Types of nodes and their functionalities adapted from authors of~\cite{colledanchise2018behavior} are briefly discussed below. 

\subsubsection{Types of nodes and functionality}
\begin{description}
\item [Sequence Node \fbox{$\rightarrow$}] can have one or more children nodes and children nodes are executed from left to right until one fails. In the context of EE dialogue modelling, a fail can refer to user wanting to interrupt the conversation while they are in the explanation strategy sub-tree to raise a disagreement and move to the disagreement sub-tree. 

\item [Priority Node \fbox{$?$}] can have one or more children nodes and children nodes are executed from left to right until one succeeds. Within an EE dialogue model, a priority node can be used in the execution of an explanation strategy behaviour tree where a list of explainers are the children nodes and they are executed based on user questions until the user indicates satisfaction. 

\begin{figure}[ht]
    \centering
    \begin{minipage}{.45\textwidth}
        \centering
        \includegraphics[width=1\linewidth]{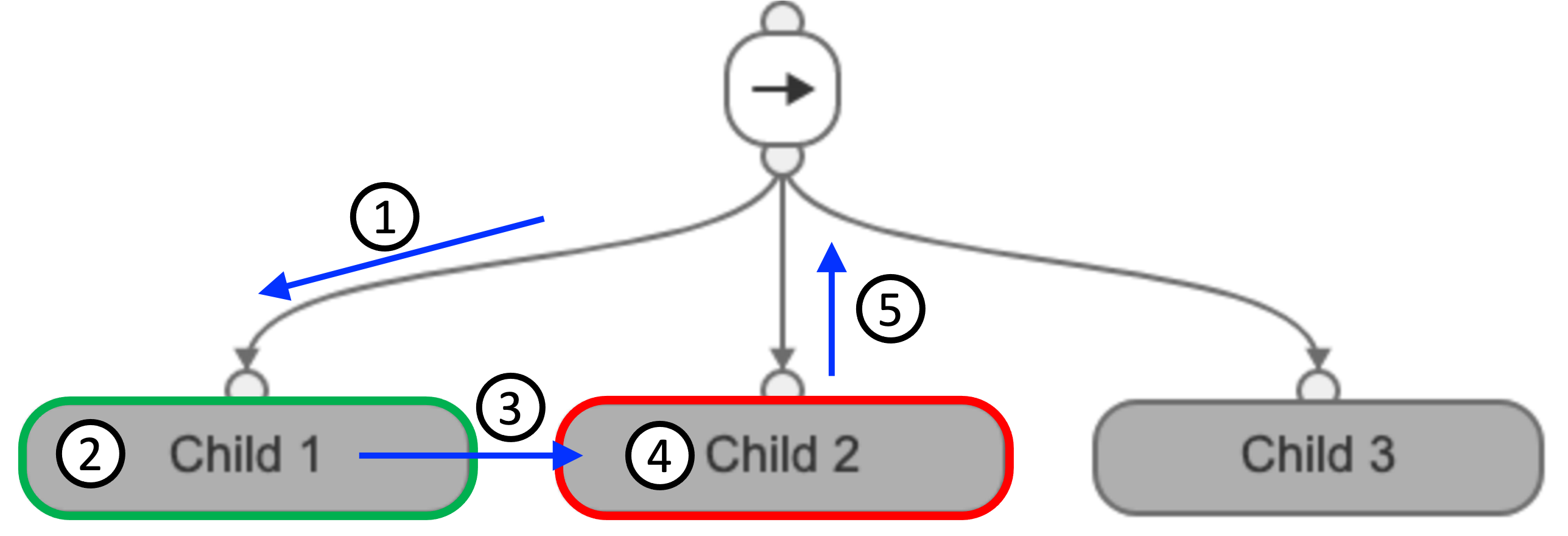}
        \caption{Sequence Node}
        \label{fig:sequence}
    \end{minipage}
    \hfill
    \begin{minipage}{0.45\textwidth}
        \centering
        \includegraphics[width=\textwidth]{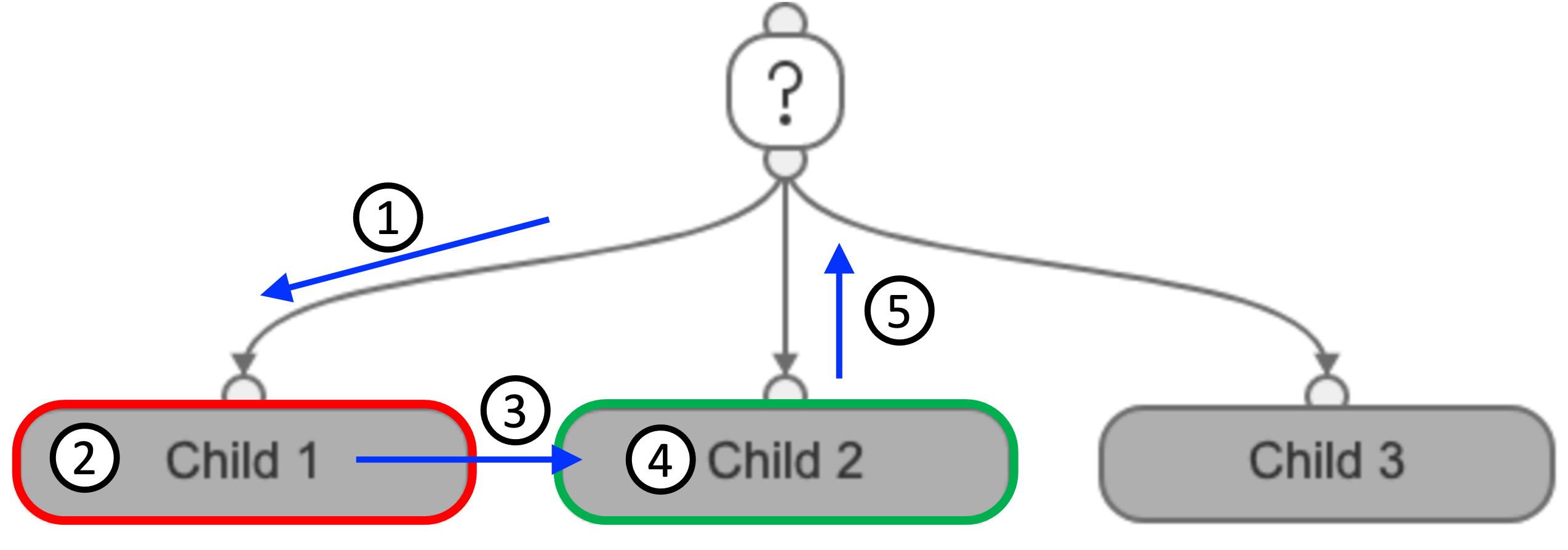}
        \caption{Priority Node}
        \label{fig:priority}
    \end{minipage}
\end{figure}

\item [Action Node] implements any behaviour. In EE dialogue modelling most commonly it will be behaviours of the chatbot in the format of \textit{the chatbot prompting the user with an utterance}, \textit{waiting for a response and analysing the response}. Based on the user response, the action node will execute its business logic to determine its status as failure or success. This will help the parent composite node~(Sequence or Priority) to decide where to navigate and which node to execute next.  

When modelling EE dialogue, we consider three subtypes of Action Nodes: Question Answer Node, Information Node and Explainer Node. Question Answer Node will pose a question to the user and wait to receive a response which decides the node status. Information Node presents information to the user without requiring a response and will always set node status to success. The Explainer Node will execute the pre-configured explainer algorithm to generate an explanations for the user. 

\item [Condition Node] performs a Boolean check, often used as the first child node of a composite node with multiple child action nodes. The Boolean check helps to control the access to all its siblings to the right. Figures~\ref{fig:condition-sequence} and~\ref{fig:condition-priority} show two scenarios where setting the $value=True$ lets us control the access to the sibling nodes. In EE dialogue, this will help to avoid repetition and improve execution efficiency. 

\begin{figure}[ht]
    \centering
    \begin{minipage}{.45\textwidth}
        \centering
        \includegraphics[width=1\linewidth]{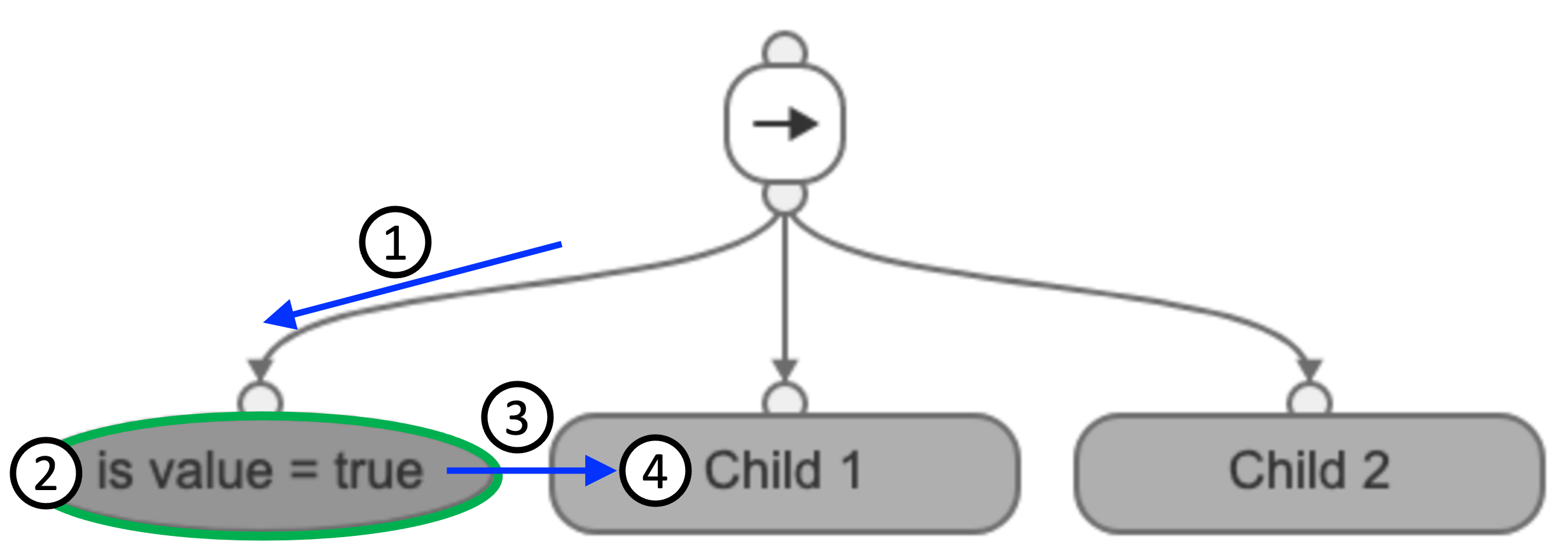}
        \caption{Condition Node used in a sequence sub-tree}
        \label{fig:condition-sequence}
    \end{minipage}
    \hfill
    \begin{minipage}{0.45\textwidth}
        \centering
        \includegraphics[width=\textwidth]{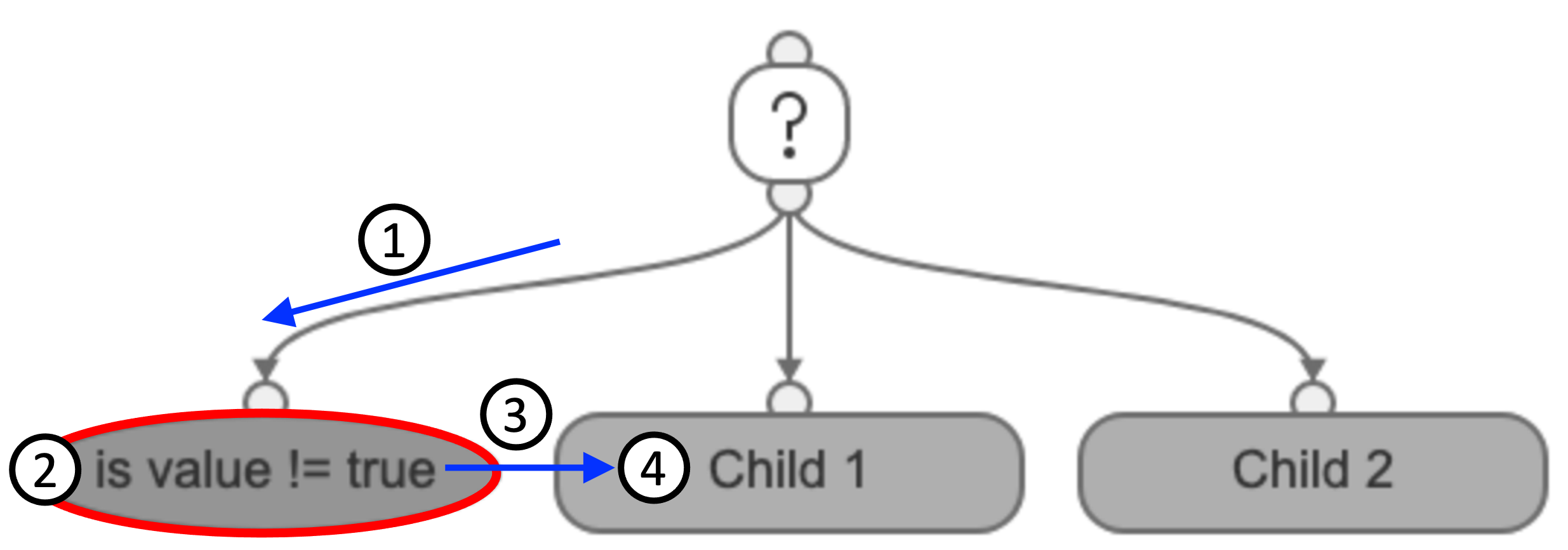}
        \caption{Condition Node used in a priority sub-tree}
        \label{fig:condition-priority}
    \end{minipage}
\end{figure}

\end{description}

\subsection{Explanation Experience Dialogue Model}
\begin{figure}[ht]
\centering
\includegraphics[width=0.9\textwidth]{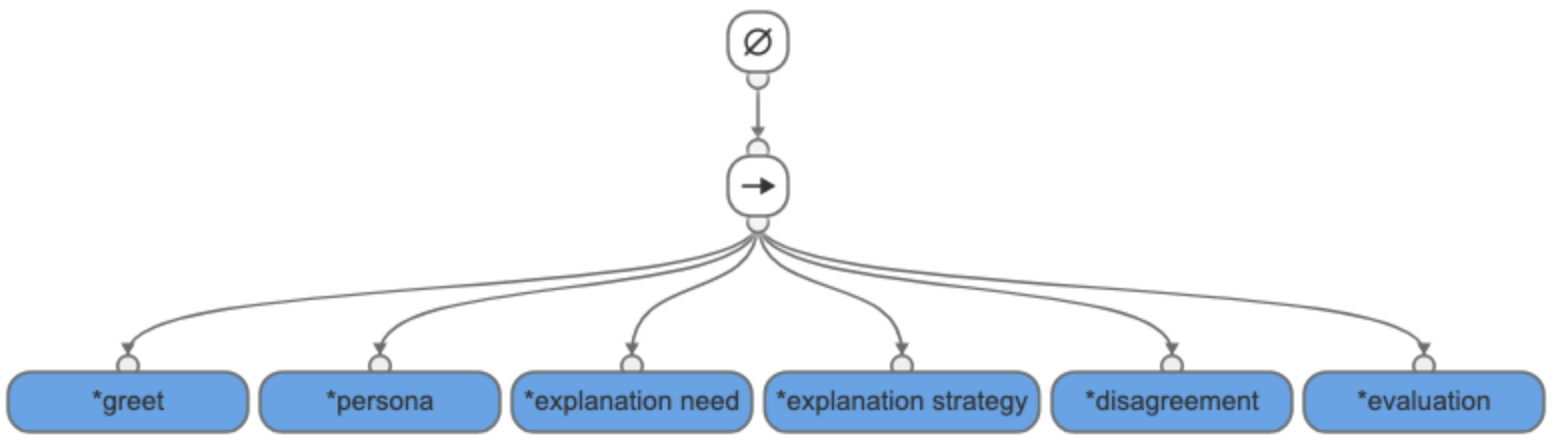}
\caption{Abstract iSee Behaviour Tree}
\label{fig:abstract_bt}
\end{figure}

An abstract BT of the EE dialogue model is presented in Figure~\ref{fig:abstract_bt}. Each child is a sub-tree that handles a specific conversational behaviour and the Asterisk indicates that all those sub-trees use a condition node as the first child to control entry. The most high-level navigation control is a sequence node which means all sub-trees should return success to complete the explanation experience, how each defines success is left to the business logic of the sub-tree. A simple execution of the chatbot would be from left to right with the following steps: 1) greet the user; 2) ask questions to establish the persona; 3) understand the explanation need by asking to select a pre-configured question or by asking their own question; 4) present the suitable explanation from the explanation strategy; 5) repeat steps~3 and~4 as needed; 6) get details and provide clarifications if a disagreement is raised; and 7) evaluate the experience using the pre-configured questionnaire. 

\subsection{Navigating the Explanation Experience Dialogue Model}
\begin{figure}[ht]
\centering
\includegraphics[width=0.9\textwidth]{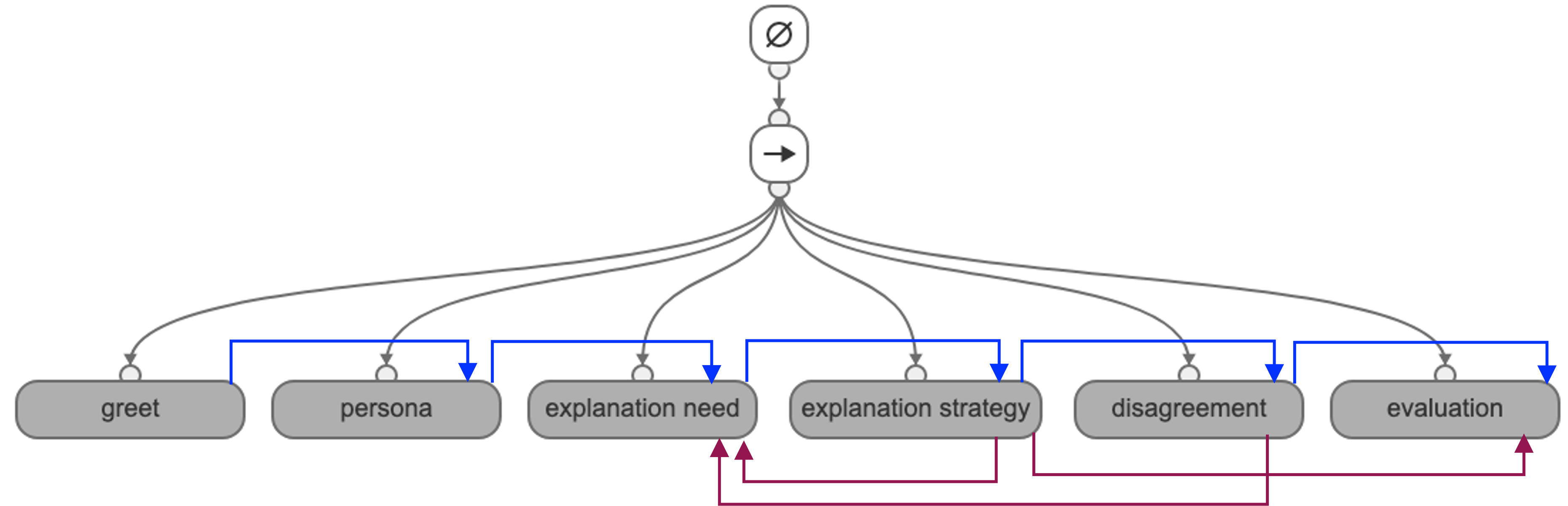}
\caption{Abstract Behaviour Tree with dependencies}
\label{fig:bt_dependencies}
\end{figure}

Figure~\ref{fig:bt_dependencies} illustrates all the navigation dependencies that are implemented as business logic in each sub-tree. Naturally, the child leaf nodes of a BT are executed from left to right; which are indicated by the blue arrows. We introduce four additional navigational dependencies to the abstract BT that are implemented in business logic using Condition Nodes, shown in red. Note that when personalising the EE dialogue model to different use cases, they can modify~(add, change, delete) navigations as required. 

\begin{description}
\item [Explanation Strategy $\rightarrow$ Explanation Need:] This enables interrupting the execution of the explanation strategy sub-tree and to invoke the explanation need sub-tree. An example scenario would be a user indicating their explanation needs have changed after receiving some explanation. Then we navigate to the explanation need sub-tree to understand their new explanation need. Once the new need is established, the natural flow of execution will take them back to the explanation strategy node to execute the relevant explanation technique. 

\item [Disagreement $\rightarrow$ Explanation Need:] If the user and the chatbot engaged in a disagreement, there are multiple scenarios where the user may need to restate their explanation need. For example, the clarification presented by the chatbot can lead the user to understand their initial explanation need is incorrect or has changed. Another example would be the user requiring additional explanations in order to support their disagreement. Accordingly, multiple nodes of the disagreement sub-tree will redirect the user to the explanation need sub-tree to restate their explanation need. 

\item [Explanation Strategy $\rightarrow$ Evaluation:] The navigation from explanation strategy to evaluation is used when user indicate their satisfaction with the explanation(s) received. Note that the user may have already had disagreements and revision of explanation needs. At this point, the chatbot directs the user to evaluate their experience which will complete their experience episode. 

\end{description}

\begin{figure}[ht]
\centering
\includegraphics[width=1\textwidth]{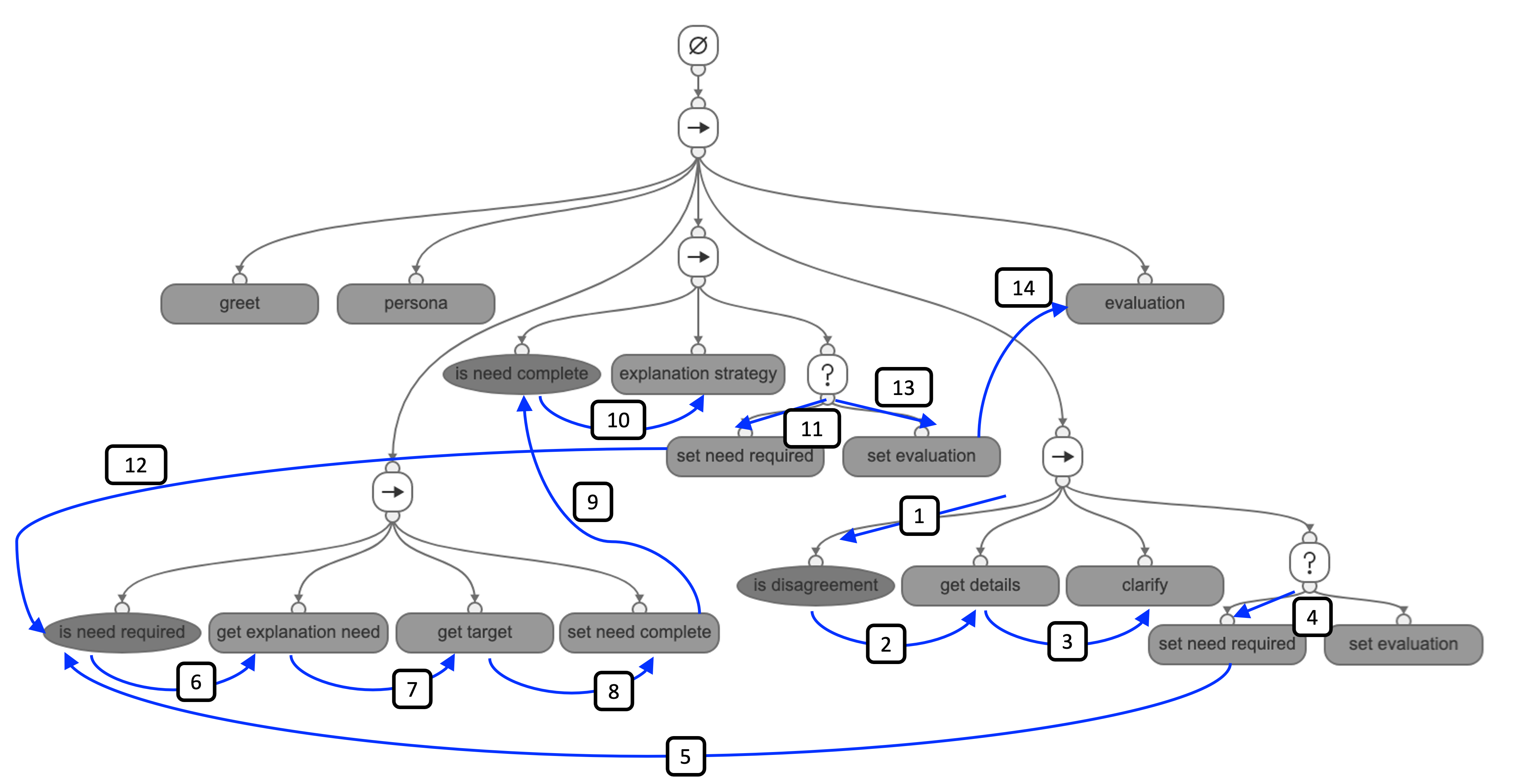}
\caption{Scenario: Disagreement $\rightarrow$ Explanation Need}
\label{fig:disagreement-need}
\end{figure}

Figure~\ref{fig:disagreement-need} illustrates a partial scenario where all three conditional navigations are used. Imagine the user and the chatbot are at a state where they enter into a disagreement~(1), the user provides details and receives clarifications~(2-3), the user indicates the need for more explanation~(4-5), the user poses a new question and provide target details~(6-9), chatbot navigates the behaviour tree and provide an explanation that answers the new question~(10), the user indicates they have other questions~(11-12) and the BT repeats steps 6 to 10~(6-10), the user indicates their satisfaction and that they are happy move to evaluation~(13), the chatbot presents the evaluation questionnaire for the user to answer~(14). Note that greet, persona and evaluation nodes are at an abstract level whereas explanation need, explanation strategy and disagreement sub-trees present fine-grained functionalities.  

\subsection{Creating Explanation Experiences}

The dialogue model is validated for robustness to dynamically updating the EE dialogue model and for creating explanation experiences for the Radiograph fracture detection use case. Firstly we used the XAI system specification from Table~\ref{tbl:jiva} to personalise the dialogue model to the Radiograph fracture detection system by reusing its Explanation Strategy in Figure~\ref{fig:jiva_bt_ano}. Apart from the explanation strategy, the persona, explanation needs and evaluation sub-trees are personalised using the information in the specification. 
Secondly, we present a conversation between a clinician and the EE chatbot which follows the conversational pathways of the personalised EE dialogue model~(in Figure~\ref{fig:jiva_bt_ano}) presented in Table~\ref{tbl:jiva_convo1}. Due to the length of the paper, we included these in Appendix~\ref{ap:convo}. 
\section{Discussion}
\label{sec:discuss}
This section evaluates the Explanation Experience~(EE) dialogue model by exploring the advantages and disadvantages of Behaviour Trees~(BT) for dialogue modelling compared to alternatives like State Transition Models~(STM). 

\subsection{Robustness to dynamic changes}

\begin{figure}[ht]
    \centering
    \begin{minipage}{.6\textwidth}
        \centering
        \includegraphics[width=1\linewidth]{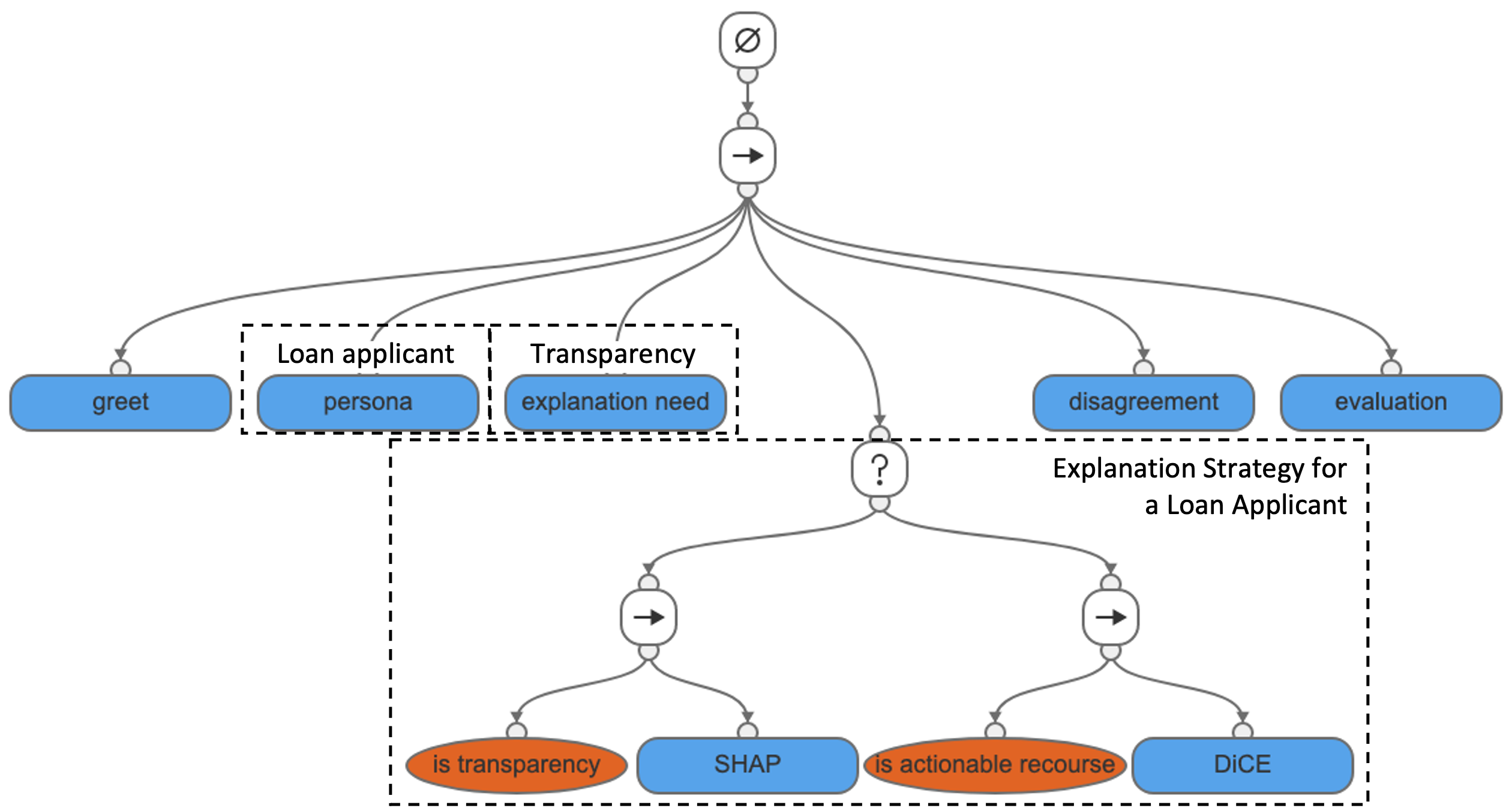}
        \caption{Loan Applicant}
        \label{fig:applicant}
    \end{minipage}
    \hfill
    \begin{minipage}{0.6\textwidth}
        \centering
        \includegraphics[width=\textwidth]{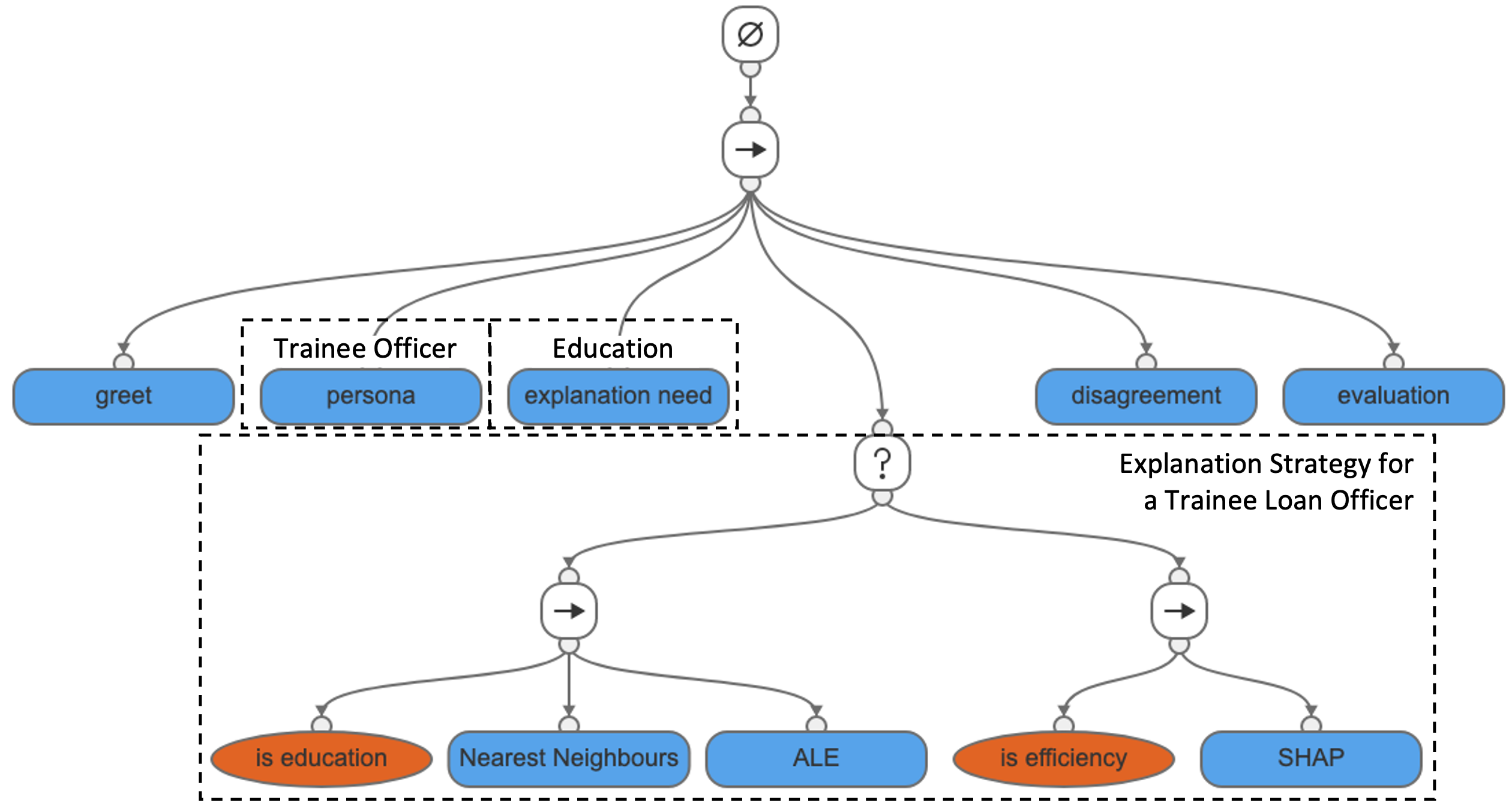}
        \caption{Trainee Loan Officer}
        \label{fig:trainee}
    \end{minipage}
\end{figure}

EE dialogue model is designed such that it is generic to any use case or persona. Each execution of the dialogue model is personalised at run-time by the XAI system specification. Accordingly, the dialogue model needs to dynamically adapt to its environment such as change of persona or explanation strategy. 
Two examples are presented in Figures~\ref{fig:applicant} and~\ref{fig:trainee} related to a loan approval XAI system. Imagine two users, a loan applicant and a trainee loan officer starting conversations with the EE chatbot. The chatbot identifies the persona based on questions about their AI and domain knowledge. This triggers a change in the dialogue model where the explanation strategy node~(in Abstract explanation strategy node in Figure~\ref{fig:abstract_bt} is now replaced by a sub-tree from the specifications for each persona~(see Table~\ref{tbl:loan} for the specification of loan applicant persona). 
BTs lend to further dynamically adapting the dialogue model in case a user raises an explanation need that is not fulfilled by the strategy designed for them in the XAI specification.

This example demonstrates BTs can be seamlessly updated without affecting existing conversational pathways. This is enabled by the ability to maintain different granularities in different parts of the BT. If a behaviour is universal across different use cases, the dialogue model can use fine-grained sub-trees. In contrast, if a behaviour is known to be dynamically changed or adapted during execution, it is left as an abstract node to be replaced by a sub-tree. 

\subsection{Reusability of components}
\begin{figure}[ht]
    \centering
    \includegraphics[width=.9\linewidth]{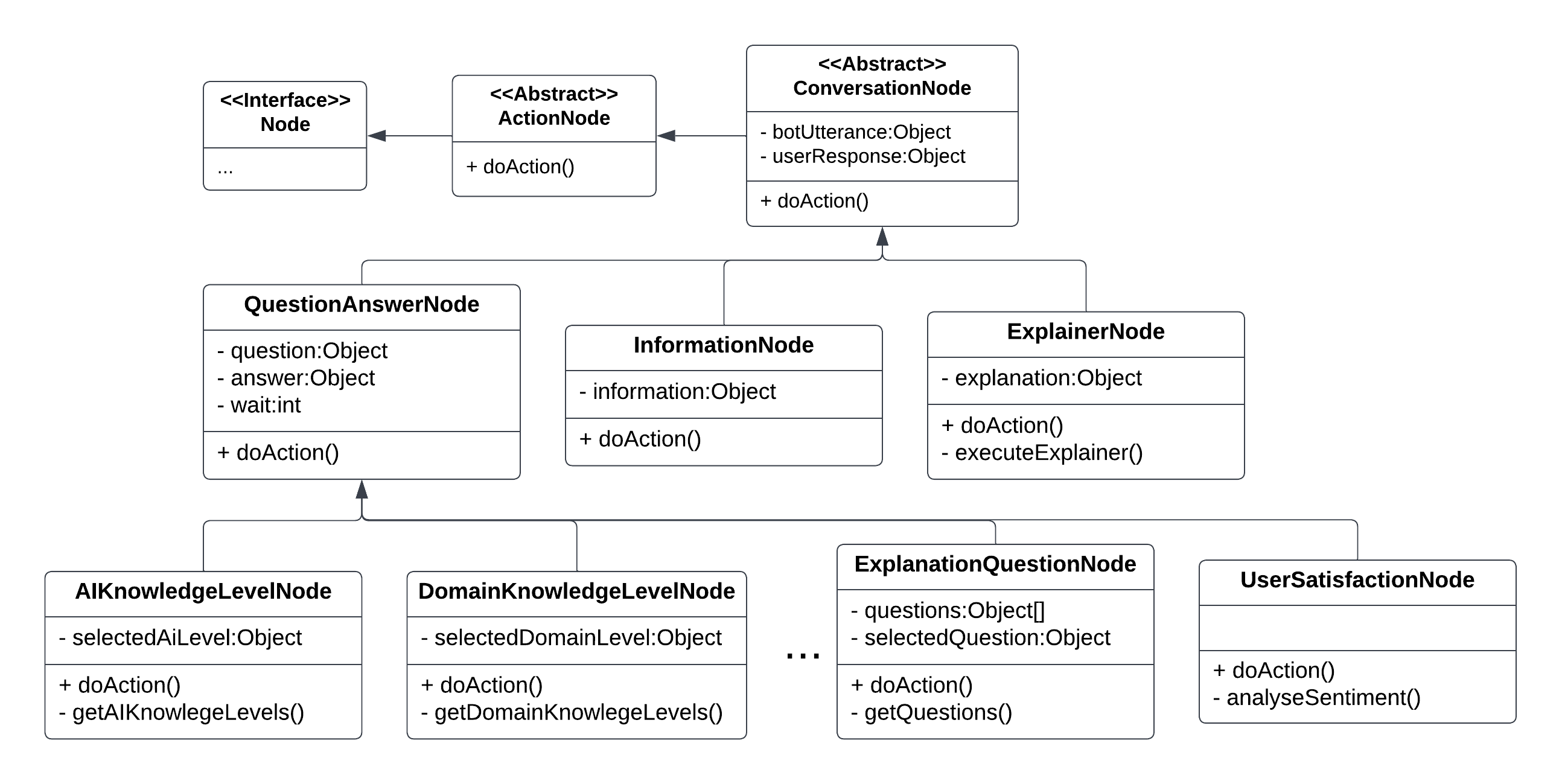}
\caption{Types of Question Answer Nodes}
    \label{fig:uml}
\end{figure}

BTs define a set of nodes that control basic navigation and behaviours~(see Section~\ref{sec:btbasics}). In order to implement the conversational behaviours required for the EE dialogue model, we introduced three new Action Nodes: Question Answer Node, Information Node and Explainer Node. Furthermore, to implement a specific conversation, we create multiple Action Nodes that inherit behaviours of one of the three Action Node types but have their own data and utterance. This property of BTs is can be seen as analogous to Object-oriented concepts of inheritance and polymorphism. Figure~\ref{fig:uml} presents the UML diagram that illustrates different types of Question Answer Nodes and the inheritance hierarchy. 

\subsection{Ability to contextualise to different output modalities}
Conversation is a well-known explanatory mechanisms~\cite{sokol2020one,miller2019explanation} which makes it the preferred modality for interactive XAI. However, there are specialised domains and user groups where the conversation maybe not be suitable. For example, for an engineer using a high-risk, time-sensitive Cybersecurity system where the AI model predicts a network attack, conversation is not the preferred modality to create EEs.
The abstract nature of BTs enables the conceptual modelling of interactions in an XAI system, without being biased towards a specific modality. In the Cybersecurity example, the behaviours of abstract nodes are still valid, however, the persona may not be captured through conversation, instead, information from the user login to the system can be used. Explanation needs may be captured through a set of rules included in the XAI system specification to determine which explanation need is applicable to a given situation. The explanation strategy can be navigated and a relevant explainer can be executed to show the explanations instantly to the user. The XAI system specification can indicate that the evaluation optional or can be postponed. 
This example shows that the execution of the EE dialogue model can be customised to the modality required by the domain. 

\subsection{Execution Cost}
The execution of BTs is significantly different to STMs. In STMs, if the conversation is at a certain state, there is a pre-defined set of possible transitions and next states. BT execution is reactive, meaning once a node is executed and decides on its status~(success or failure) any other node can execute given the navigational constraints are satisfied. These constraints can take the form of composite nodes or condition nodes controlled by business logic. Accordingly, all other ``candidate'' nodes continuously ``listen and wait'' to be executed. 

This can be computationally expensive when executing a BT. However, there are multiple methods to improve execution efficiency by minimising the number of ``candidate'' nodes waiting. In EE dialogue model we opt for using \textit{memory} implemented with Condition Nodes. The access to a sub-tree with multiple action nodes is controlled by using a condition node as the left-most sibling. As a result, only the condition node needs to ``listen and wait'' instead of all action nodes in the sub-tree. There is still a significant computational cost in comparison to executing an STM. However, the reactive nature of BTs lends to capturing rich EEs by creating close-to-natural conversational pathways.

\subsection{Ability to control complexity}
EE dialogue model is abstract such that it is applicable to any XAI system. Given an XAI system, the specification contains information to personalise a node~(persona question about AI Knowledge level) or a sub-tree~(explanation strategy). This information is curated by a set of individuals including XAI, domain and system experts who are not necessarily experts in dialogue design. Accordingly, the conceptual model should be interpretable to everyone involved in the design. 
The ability to maintain sub-trees in different granularities lends well to creating BTs that are interpretable to humans. The coarse granular sub-trees hide the complexities from a lay person. The XAI expert can focus on the fine-grained design of the explanation strategy sub-tree, and domain and system experts can validate the automatically generated evaluation sub-tree from their questionnaire.  Accordingly, the modularity of the BTs enables to hide the complexities from different audiences as necessary. 

\subsection{Summary}
BTs contribute to a dialogue model that is robust yet open compared to a STM. The robustness to dynamic change in the behaviour model, reusability of behaviours, ability to contextualise to different output modalities, and controlled complexity for interpretability to a larger audiance are four main benefits we highlight for choosing BT over STMs for the EE dialogue model. There is an increased computational complexity which we view as an intrinsic cost for maintaining open conversational pathways. In addition to the aforementioned benefits, we inherit some native properties of BTs over other conceptual models such as less susceptibility to an error in implementation, a reduced number of nodes compared to other tools like Decision Trees and having multiple entry points to conversation compared to a single entry point in STMs.  

\section{Conclusion}
\label{sec:conc}
Interactive multi-shot explanation experiences are key to achieving user satisfaction and improved trust in using AI for decision-making. This paper presented an XAI system specification and an interactive dialogue model to facilitate the creation of these explanation experiences. Such specifications combine the knowledge of XAI, domain and system experts of a use case to formalise target user groups and their explanation needs and to implement explanation strategies to address those needs. Formalising the XAI system promotes the reuse of existing explainers and known explanation needs that can be refined and evolved over time using evaluation feedback from the users. The abstract EE dialogue model formalised the interactions between a user and an XAI system. The resulting EE conversational chatbot is personalised to an XAI system at run-time using the knowledge captured in its XAI system specification. 
This seamless integration is enabled by using BTs to conceptualise both the EE dialogue model and the explanation strategies. 
In the evaluation, we discussed several desirable properties of using BTs over traditionally used STMs or FSMs. BTs promote reusability of dialogue components through the hierarchical nature of the design.   
Sub-trees are modular, i.e. a sub-tree is responsible for a specific behaviour, which can be designed in different levels of granularity to improve human interpretability.
The EE dialogue model consists of abstract behaviours needed to capture EE, accordingly, it can be implemented as a conversational, graphical or text-based interface which caters to different domains and users. 
There is a significant computational cost when using BTs for modelling dialogue, which we mitigate by using memory. Overall, we find that the ability to dynamically create robust conversational pathways makes BTs a good candidate for designing and implementing conversation for creating explanation experiences.

\bibliographystyle{ieeetr}
\bibliography{ref}

\clearpage
\appendix
\section{XAI System Specifications}
\label{ap:xaispecs}

\begin{landscape}
\begin{table}[ht]
\centering
\caption{XAI System: For Loan applicants using the Loan approval system}
\small
\renewcommand{\arraystretch}{1.2}
\begin{tabular} {lp{14cm}}
\hline
\textbf{System Description}\\
\textit{\tab AI System}\\
\tab \tab AI Task&Given a loan application, the system predicts if it is "Approved" or "Rejected"\\
\tab \tab AI Method&A Random Forest performing binary classification\\
\tab \tab Data&AI Model was trained with 342865 data instances, each with 69 attributes that describe a loan application, some attributes are categorical and others are numeric. \\
\tab \tab Assessment& Accuracy of the model is 99\%\\
\textit{\tab Persona: loan applicant}\\
\tab \tab AI Knowledge Level&no knowledge\\
\tab \tab Domain Knowledge Level&Novice\\
\tab \tab Resources&Screen Display\\
\textit{\tab Explanation Need}\\
\tab \tab User questions and Intents&Why was my loan application rejected? $\rightarrow$ Transparency intent \\
&What changes would get my loan application approved? $\rightarrow$ Actionable recourse intent\\
\tab \tab Target&A loan application characterised by 69 features: Loan amount: 10000, funded amount: 10000, term: 36 months, interest rate: 13.5, instalment: 343.3, home ownership: rent,  ... with outcome either "Approved" or "Rejected"\\
\hline
\textbf{Explanation Strategy}\\
\textit{\tab Explainers}&Feature attribution using SHAP to satisfy transparency intent\\
&Counterfactual explanation using DiCE to satisfy actionable recourse intent\\
\textit{\tab Behaviour Tree}&\includegraphics[width=.5\textwidth]{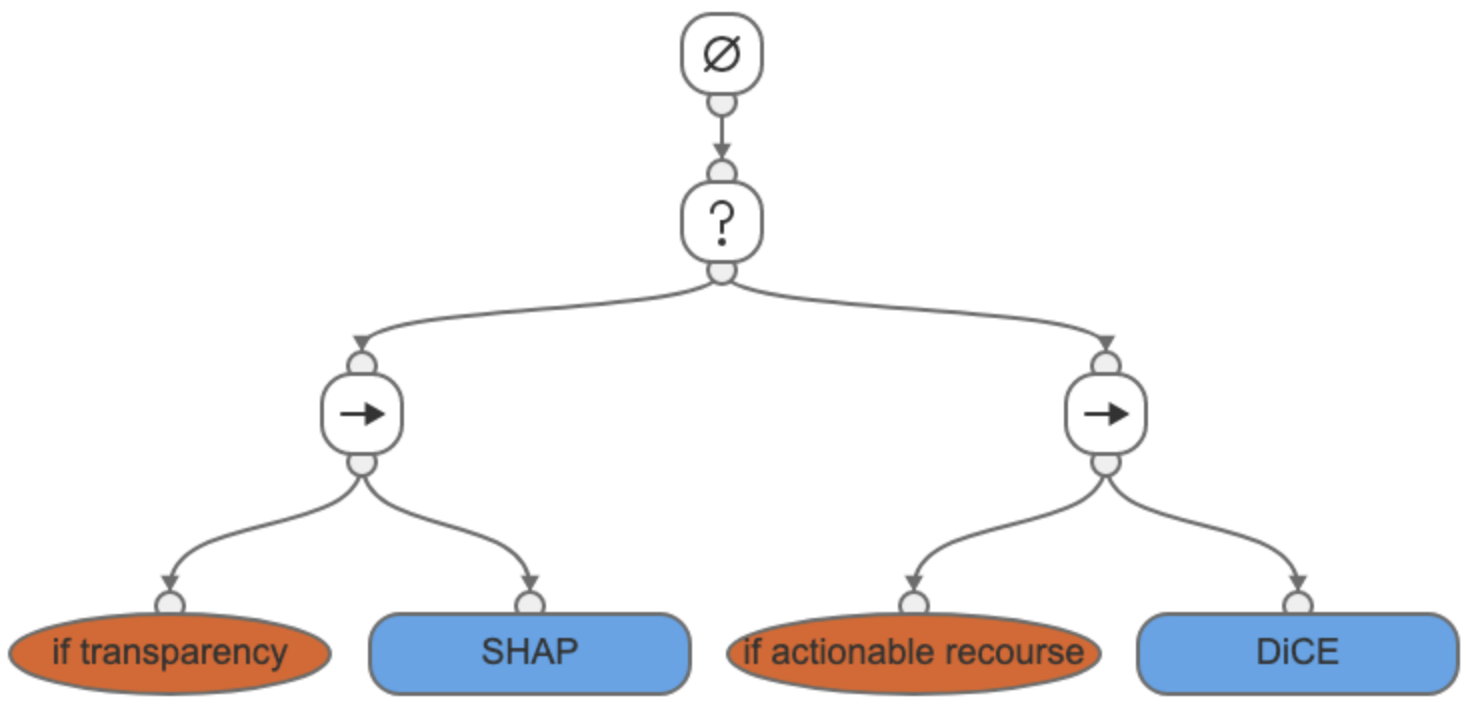}\\
\hline 
\end{tabular}
\label{tbl:loan}
\end{table}
\end{landscape}

\begin{landscape}
\begin{table}[ht]
\centering
\small
\renewcommand{\arraystretch}{1.2}
\begin{tabular} {lp{14cm}}
\hline
\textit{\tab Behaviour Tree Description}& If the user has transparency intent, present a feature attribution explanation using LIME, else if the user has actionable recourse intent, present a counterfactual using DiCE. If the user is satisfied after receiving an explanation for one intent exit, else, present the remaining explanation. Exit after executing both explanation techniques~(regardless of user satisfaction to avoid repetition)\\
\hline
\textbf{Evaluation Strategy}\\
\textit{\tab Questionnaire}&Question 1: Now I have a better understanding of how the system works.\par Answers: Likert scale, 5 items from Strongly Disagree to Strongly Agree\\
&Question 2: The explanations that were presented had sufficient detail.\par Answers: Likert scale, 3 items from Disagree to Agree\\
&Question 3: This experience helps me judge when I should trust and not trust the system.\par Answers: Yes or No\\
\textit{\tab Interpretation policy}&2 out of the 3 questions should receive positive responses from the group of loan applicants\\
\hline
\end{tabular}
\end{table}
\end{landscape}

\begin{landscape}
\begin{table}[t]
\centering
\caption{XAI System: For judges using the recidivism prediction system}
\footnotesize
\renewcommand{\arraystretch}{1.2}
\begin{tabular} {lp{14cm}}
\hline
\textbf{System Description}\\
\textit{\tab AI System}\\
\tab \tab AI Task&System predicts recidivism within 2 years of an inmate based on demographic information and prior record\\
\tab \tab AI Method&A RandomForest performing multi-class classification\\
\tab \tab Data&AI Model was trained with 18610 data instances, each data instance is an inmate in the system between 2013 to 2014, described using attributes like age, race, sex and priors count\\
\tab \tab Assessment&Accuracy of the AI system is 63.6\%\\
\textit{\tab Persona: Judge}\\
\tab \tab AI Knowledge Level&no knowledge\\
\tab \tab Domain Knowledge Level&expert\\
\tab \tab Resources&Screen Display\\
\textit{\tab Explanation Need}\\
\tab \tab User questions&Why does the system predict a high risk of recidivism for the inmate? $\rightarrow$ Education intent\\
&In general, which attributes contribute most to predicting high risk in the AI system? $\rightarrow$ Transparency intent\\
&If I ignore race and sex, does the system prediction change? $\rightarrow$ Scrutability intent\\
\tab \tab Query&Inmate profile for which an explanation is needed, the AI model with outcome either High, Medium or Low risk\\
\hline
\textbf{Explanation Strategy}\\
\textit{\tab Explanation techniques}&Nearest-neighbour to satisfy education intent\\
&SHAP to satisfy transparency intent\\
&TwinCBR and ALE to satisfy scrutability intent\\

\hline 
\end{tabular}
\label{tbl:recidivism}
\end{table}
\end{landscape}

\begin{landscape}
\begin{table}[ht]
\centering
\small
\renewcommand{\arraystretch}{1.2}
\begin{tabular} {lp{14cm}}
\hline
\textit{\tab Behaviour Tree}&\includegraphics[width=.7\textwidth]{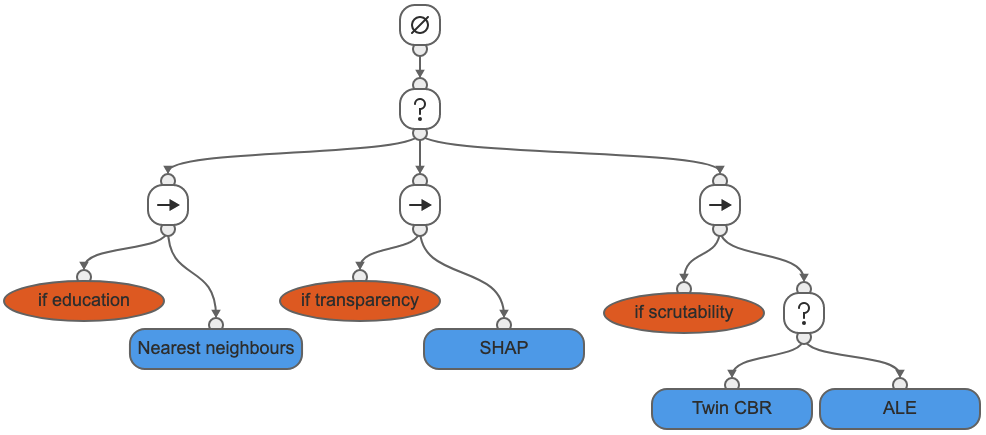}\\
\textit{\tab Behaviour Tree Description}&If the user indicates education intent, present 3 nearest neighbour inmates similar to the query inmate; if the user indicates transparency intent, present feature attributions using SHAP explanation technique and if the user indicates scruitability intent, first create a CBR Twin to match their requirements and present an explanation, if the user is not satisfied, present global ALE for each attribute. In case the user indicates satisfaction after receiving an explanation, exit, or else present explanations for remaining intents. Exit after executing all four explanation techniques.\\
\hline
\textbf{Evaluation Strategy}\\
\textit{\tab Questionnaire}&Question 1: The system provides sufficient details when explaining predictions.\par Answers: Likert scale, 5 items from Strongly Disagree to Strongly Agree\\
&Question 2: I like using the system for decision-making in my work.\par Answers: Likert scale, 5 items from Strongly Disagree to Strongly Agree\\
&Question 3: This experience helps me judge when I should trust and not trust the system.\par Answers: Likert scale, 5 items from Strongly Disagree to Strongly Agree\\
\textit{\tab Interpretation Policy}&all questions should receive positive responses from the users\\
\hline
\end{tabular}
\end{table}
\end{landscape}
\clearpage
\section{An Explanation Experience: Clinician using Radiograph fracture detection XAI system Chatbot}
\label{ap:convo}

\begin{figure}[ht]
\centering
\includegraphics[width=.9\textwidth]{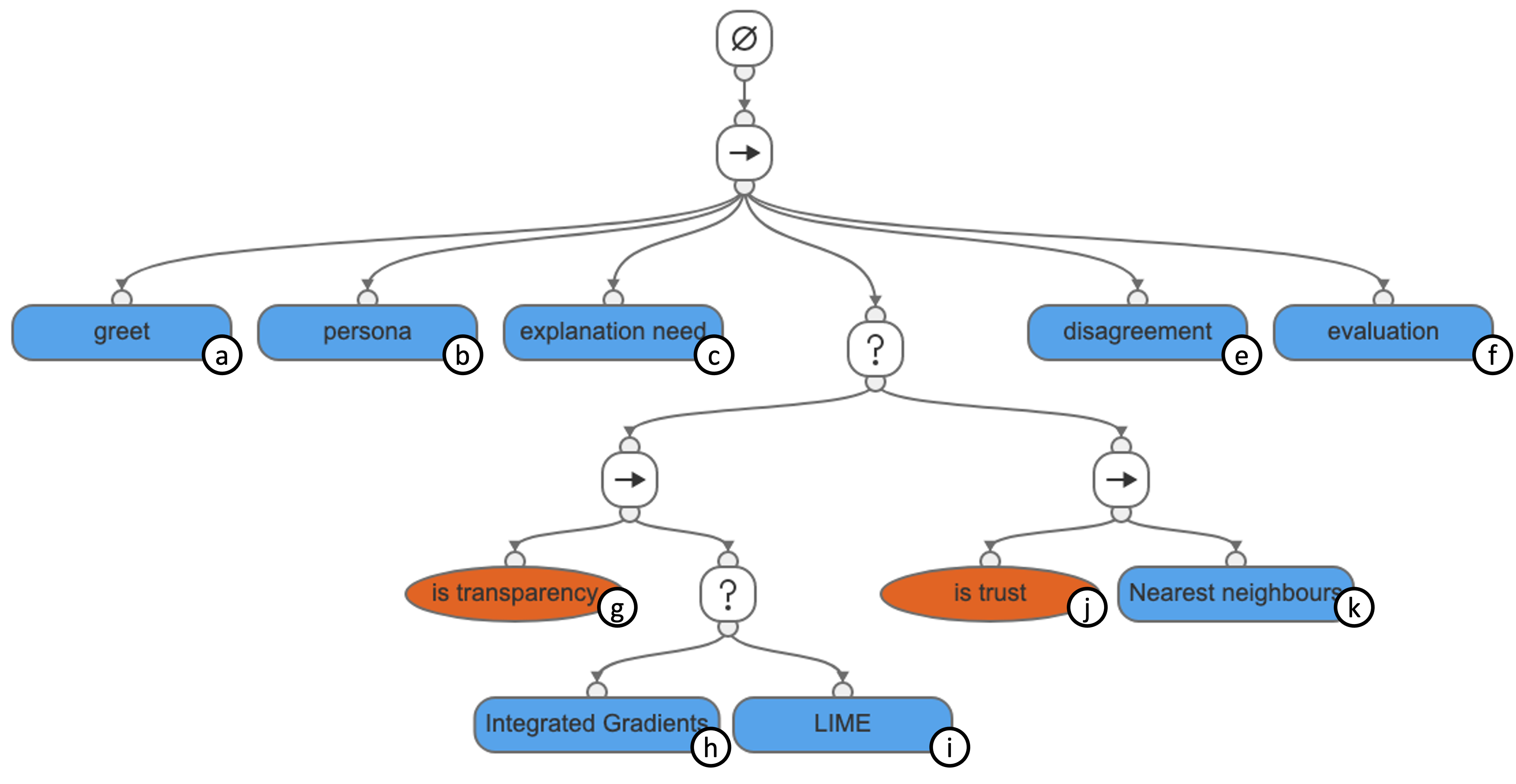}
\caption{Annotated Behaviour Tree for Radiograph Fracture Detection XAI System}
\label{fig:jiva_bt_ano}
\end{figure}

\begin{landscape}
\begin{table}[t]
\centering
\caption{Explanation Experience of a clinician using the EE Chatbot}
\footnotesize
\renewcommand{\arraystretch}{1.1}
\begin{tabular} {llp{14cm}l}
\hline
Node&Actor&Utterance&Status\\
\hline
\fbox{a}&bot&Hello! I am the EE chatbot for the Radiograph Fracture Detection System. First I need to ask few questions to establish your persona. Would you like to proceed?\\
&user&Yes of course!&Success\\
\hline
\fbox{b}&bot&What is your level of knowledge on AI?\\
&user&I have no understanding of AI technology.\\
&bot&What is your level of knowledge in the domain of fracture detection?\\
&user&I have been a practising clinician for 12 years. So I would say I am very knowledgeable\\
&bot&Thank you for answering the questions.&Success\\
\hline
\fbox{c}&bot&Next I want to understand what kind of explanation you want. Please select a question below if it is similar to what you would like to know, or tell me what you would like to know.\par
1. Why is this Radiograph marked as ”fracture”?\par
2. Show me similar Radiographs that are also marked as ”fracture”?\\
&user&Question 2 sounds like what I need to know about this specific Radiograph.\\
&bot&Thanks. Can you confirm this is the Radiograph for which you need an explanation? and the outcome you received is that it contains a fraction?\par
\includegraphics[width=.16\textwidth]{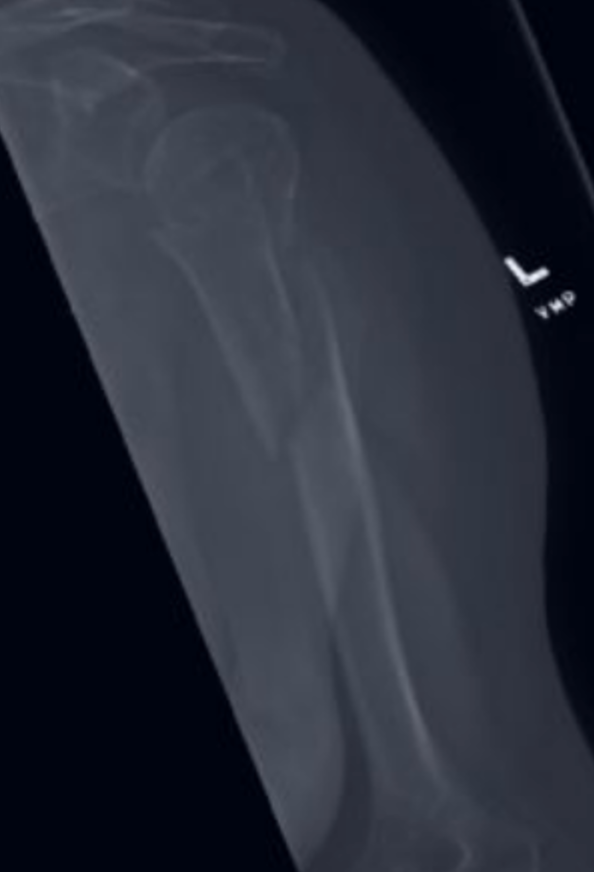}\\
&user&yes this is correct!\\
\hline
\end{tabular}
\label{tbl:jiva_convo1}
\end{table}
\end{landscape}

\begin{landscape}
\begin{table}[t]
\centering
\footnotesize
\renewcommand{\arraystretch}{1.1}
\begin{tabular} {llp{14cm}l}
\hline
&bot&Thanks, Let me find an explanation for you.&Success\\
\hline
\fbox{j}$\rightarrow$\fbox{k}&bot&Here are the two other Radiographs from our database that are most similar to your Radiograph. \par
\includegraphics[width=.32\textwidth]{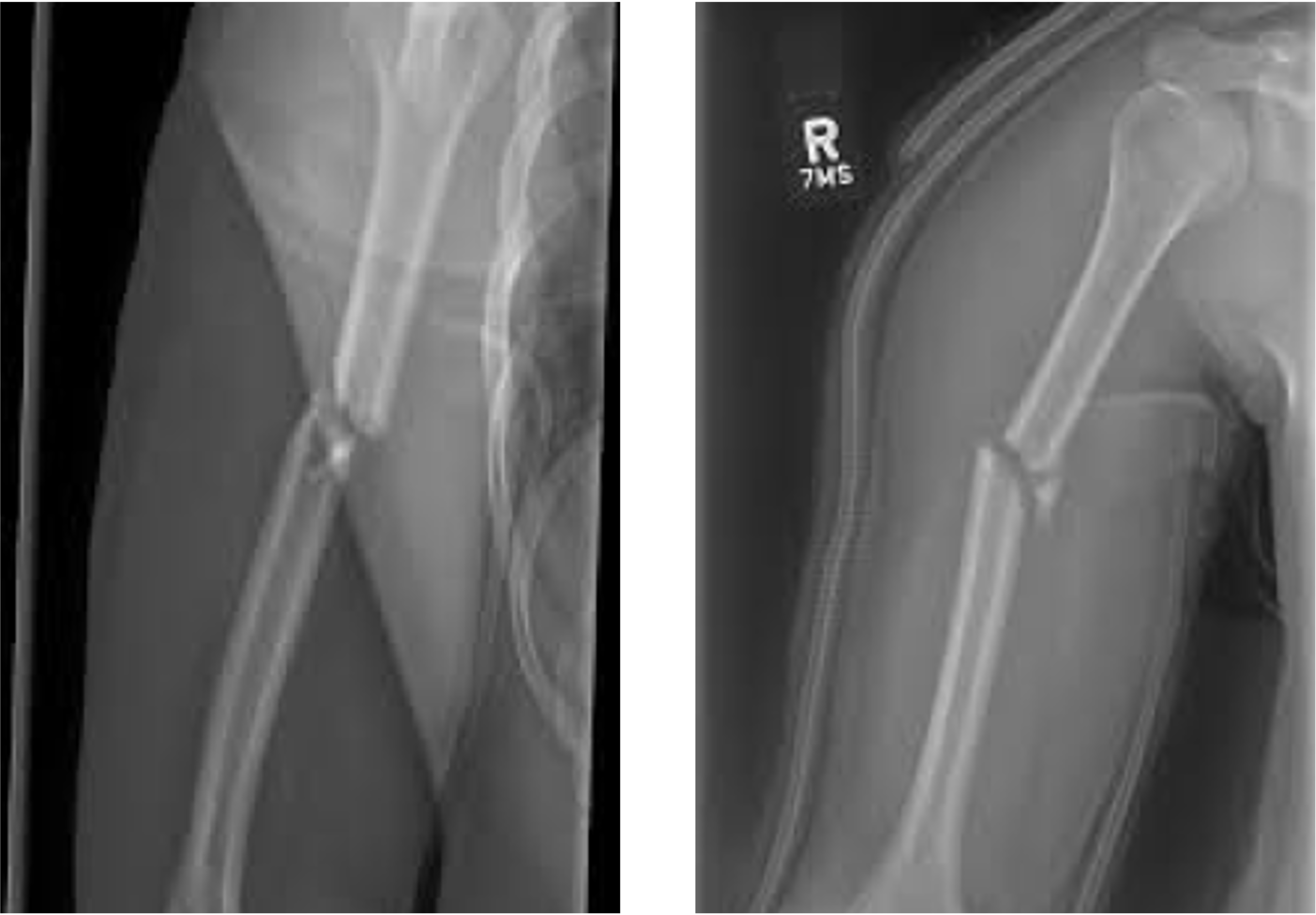}\par
What do you think?\\
&user&Okay. I see why the system thinks this is a fracture. What else can you tell me about this Radiograph?&Fail\\
\hline
\fbox{g}$\rightarrow$\fbox{h}&bot&I can also show you exact areas of the Radiograph let the system to identify the fracture
\par
\includegraphics[width=.16\textwidth]{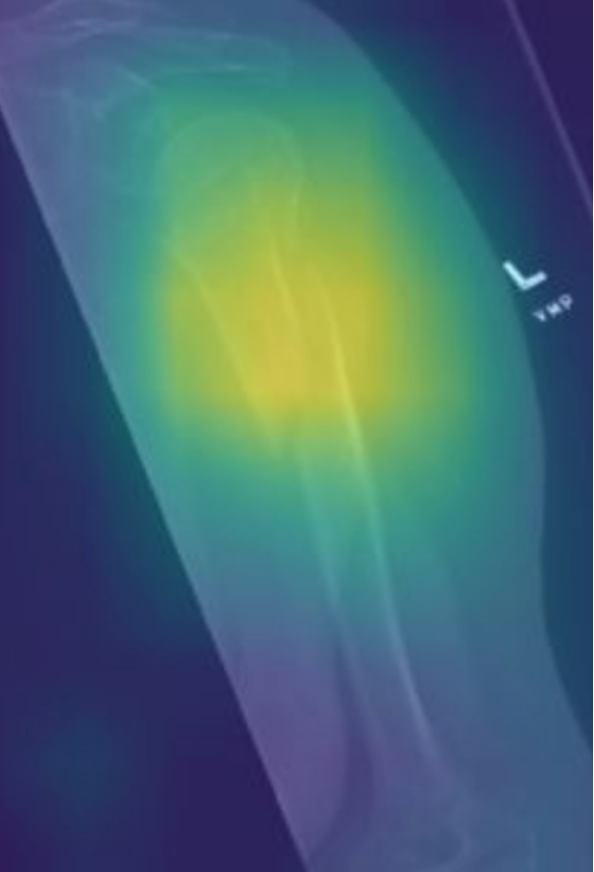}\par
Do you think we got it right?\\
&user&I'm not sure I agree&Fail\\
\hline
\fbox{e}&bot&I see... can you tell me a bit more about why you think so?\\
&user&well if you look closely, there is also a hairline fracture at the bottom left corner the system missed\\
&bot&Thank you for that information. At the moment the system is correct 83\% of the time. We will use your feedback to improve the system. \\
&user&Okay!&Success\\
\hline
\end{tabular}
\end{table}
\end{landscape}

\begin{landscape}
\begin{table}[t]
\centering
\footnotesize
\renewcommand{\arraystretch}{1.1}
\begin{tabular} {llp{14cm}l}
\hline
\fbox{f}&bot&Anything else I can help with you today? Or would you like to take a few questions to evaluate your experience?\\
&user&Can I see two more similar Radiographs?&Fail\\
\hline
\fbox{j}$\rightarrow$\fbox{k}&bot&Certainly. Here are two other similar Radiographs. These are not as similar to the ones I showed earlier.\par \includegraphics[width=.32\textwidth]{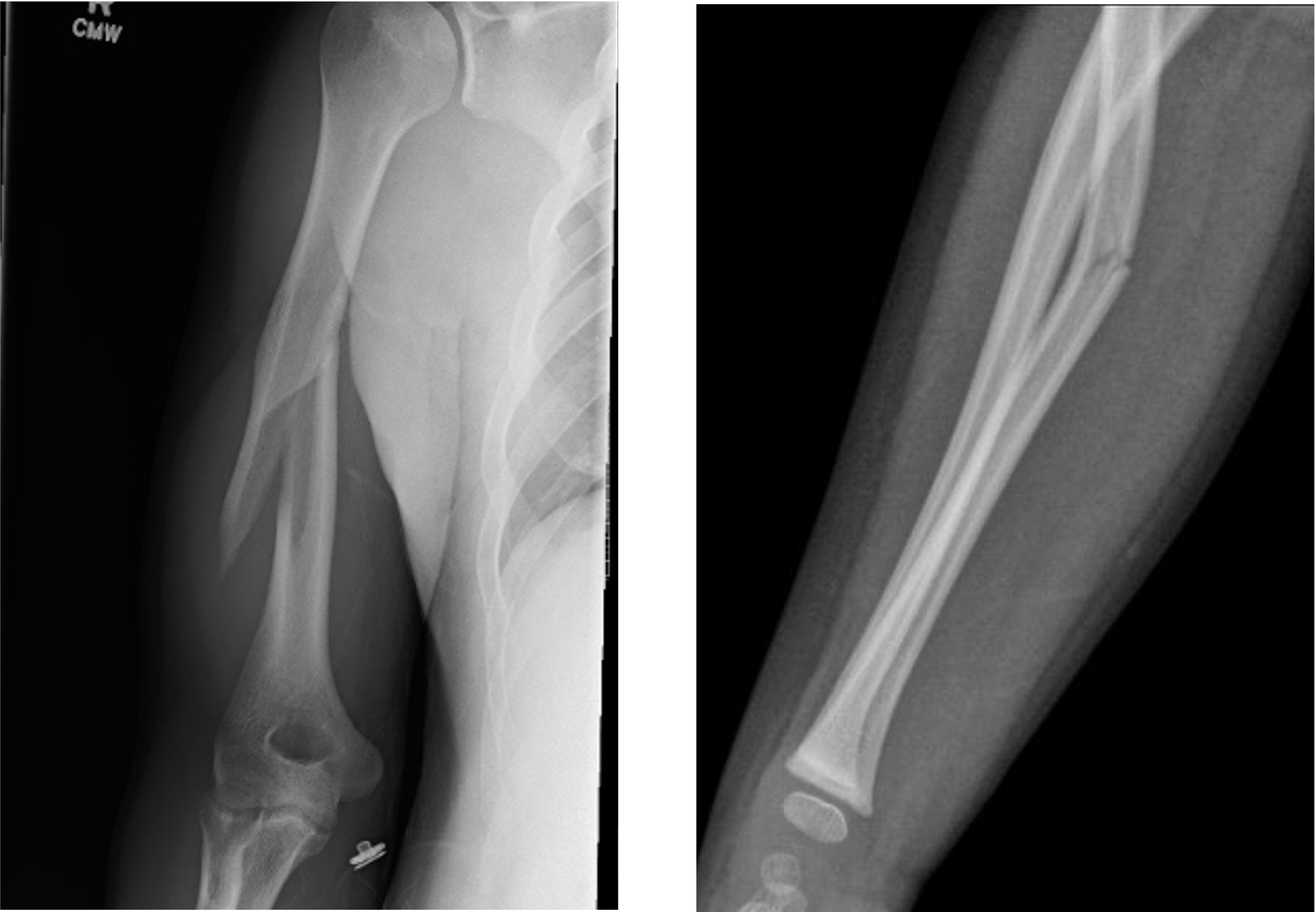}\\
&user&Okay. Thanks!&Success\\
\hline
\fbox{f}&bot&would you like to take the questionnaire now?\\
&user&Sure\\
&bot& I have three statements, for each one please answer with a response from the following. Strongly agree, Agree, Neutral, Disagree, Strongly Disagree\\
&bot&Statement 1: The explanations that were presented had sufficient detail.\\
&user&Agree\\
&bot&Statement 2: The explanations let me know how accurate or reliable the system is.\\
&user&Neutral\\
&bot&Statement 3: The explanation lets me know how trustworthy the system is.\\
&user&Neutral\\
&bot&Thank you for your feedback. Have a nice day!&Success\\
\hline
\end{tabular}
\end{table}
\end{landscape}

\end{document}